%% file: main.tex
\pdfoutput=1
\documentclass{article}
\usepackage{iclr2027_conference,times}
\iclrfinalcopy

\usepackage{latexsym}
\usepackage[T1]{fontenc}
\usepackage[utf8]{inputenc}
\usepackage{microtype}
\usepackage{inconsolata}
\usepackage{booktabs}
\usepackage{graphicx}
\usepackage{float}   %
\usepackage{placeins}  %
\usepackage{amsmath}
\usepackage{amssymb}
\usepackage{xspace}
\usepackage{array}
\input{sections/table_typography.tex}
\usepackage{multirow}
\usepackage{xcolor}
\definecolor{fgreen}{HTML}{117733}
\definecolor{fred}{HTML}{CC3311}
\definecolor{fblue}{HTML}{0072B2}
\definecolor{revblue}{HTML}{1F4FB4}
\newif\ifrevmark\revmarkfalse  %
\newcommand{\rev}[1]{\ifrevmark\textcolor{revblue}{#1}\else#1\fi}
\newcommand{\revon}{\ifrevmark\color{revblue}\fi}
\makeatletter  %
\@namedef{revbib@bullenkamp2026synthetic}{}\@namedef{revbib@mara2019heicubeda}{}
\let\rev@bibitem\bibitem
\def\bibitem[#1]#2{\rev@bibitem[#1]{#2}\ifrevmark\@ifundefined{revbib@#2}{\color{black}}{\color{revblue}}\fi}
\makeatother
\usepackage{tikz}
\usetikzlibrary{arrows.meta,positioning,calc,fit,backgrounds}
\usepackage{url}
\usepackage{xurl}
\usepackage{hyperref}

\newcommand{\vlm}{\textsc{PaddleOCR-VL}\xspace}
\newcommand{\crnn}{\textsc{CRNN}\xspace}
\newcommand{\svtr}{SVTRv2\textsuperscript{*}\xspace}

\newcommand{\lcs}{LCS Recall\xspace}

\newcommand{\SHUFSEEDS}{We trained a second Stage-A model of each kind (seed 123, same corpora) and evaluated all four at $N$=0. \lcs (\%) for real-order / shuffled models, at seeds 42 and 123, is: in-period 66.8/62.5 and 66.9/63.2 ($\Delta$ = $+$4.3, $+$3.7); cross-period 59.7/61.1 and 61.2/61.0 ($-$1.4, $+$0.2); cross-collection 30.7/33.2 and 32.9/32.6 ($-$2.4, $+$0.2). Real-order seeds differ by 0.1/1.5/2.1\,pp across the three sets; shuffled seeds differ by 0.7/0.1/0.5\,pp. The in-period advantage is stable and larger than the observed seed variation. Off-domain effects are smaller and seed-dependent. This global shuffle is stronger than the primary within-split shuffle, which changes co-occurrence and order jointly. The primary study finds \rev{an} all-lines $N$=100 SER benefit \rev{(Table~\ref{tab:ladder_panelb}) that is positive in-period for all three Stage-A seeds (Appendix~\ref{app:stageaseeds}).} These are complementary sensitivity analyses. Stage-B CIs in Table~\ref{tab:shuf} condition on the seed-42 Stage-A checkpoints.}
\newcommand{\VLMTEXTEXPOSURE}{The VLM's cells move similarly ($N$=25/100: 73.7/83.4 text-exposed vs.\ 74.7/82.7 text-disjoint); its $N$=0 change is $-$2.0\,pp.}

\definecolor{hlblue}{rgb}{0.12,0.47,0.71}
\hypersetup{colorlinks=true,linkcolor=hlblue,citecolor=hlblue,urlcolor=hlblue}

\title{From Signs to Sequences: Compositional\\
       Synthesis for Cuneiform Recognition\\
       Across Periods and Collections}

\author{Tiantong Lu\\
{\normalfont University of California, Berkeley}\\
{\normalfont\small lutt@berkeley.edu}
\And Jerry Cheng\\
{\normalfont University of California, Berkeley}\\
{\normalfont\small jerrycheng03@berkeley.edu}
\AND Adam G. Anderson$^{\dagger}$\\
{\normalfont Research Director, FactGrid Cuneiform Project}\\
{\normalfont\small adam.anderson@factgrid.eu}}

\hypersetup{pdftitle={From Signs to Sequences: Compositional Synthesis for Cuneiform Recognition Across Periods and Collections},pdfauthor={Tiantong Lu, Jerry Cheng, Adam G. Anderson}}
\begin{document}
\maketitle
\lhead{Preprint}
{\renewcommand{\thefootnote}{\fnsymbol{footnote}}\footnotetext[2]{Corresponding author.}}

\begin{abstract}
Training data that pair cuneiform line images with sign sequences are scarce, even though
isolated sign images and machine-readable texts are available separately. This scarcity
keeps large collections of digitized tablets difficult to search and analyze automatically.
We introduce a compositional supervision method that combines these independently available
resources. Given a normalized sign sequence from a text corpus, we sample corresponding real
sign crops from a tablet-disjoint visual dictionary (augmented for rare signs) and compose
them into a synthetic line image. These image--sequence pairs pre-train three recognition
architectures---a CRNN, an SVTRv2-inspired CTC model, and PaddleOCR-VL---before adaptation
on a limited number of annotated tablets. In the principal evaluation, sign locations are
supplied by ground-truth annotations and the input lines are reconstructed from their crops.
With a fixed sign inventory, adaptation on 100 Ur~III tablets yields sign error rates (SER)
of 18--25\% on in-period data. Removing compositional pre-training increases error by
24--52 percentage points across architectures and evaluation domains. \rev{On natural line images
cut from the tablet without stitching, compositional pre-training lowers the SVTRv2-inspired
model's in-period SER from 89.2\% to 52.8\%.} Controlled
experiments within the SVTRv2-inspired model further show that multi-sign sequence training
provides substantially greater benefits than isolated-sign training, whereas preserving
attested sign order and co-occurrence produces smaller, domain-dependent gains. Adaptation
primarily improves recognition within the target period, while cross-period and
cross-collection transfer remains difficult. \rev{A} modular pipeline with predicted
sign locations and reading order reaches 38.05\% whole-side SER on reused in-period
tablets, versus 68.92\% for its strongest original geometric \rev{baseline; its compositionally
pre-trained sequence verifier also lowers SER on source-disjoint tablets.} These results
establish a practical way to repurpose existing Assyriological image and text collections
as sequence-level supervision, while clarifying the remaining challenges for scalable
cuneiform image-to-text systems.
\end{abstract}

\section{Introduction}
\label{sec:intro}

Cuneiform tablets are being digitized at scale in repositories such as the Cuneiform Digital Library Initiative (CDLI)
\citep{cdli2023} and the Electronic Babylonian Library (eBL) \citep{che2026ebl}, and
automated reading could accelerate scholarly access. Annotated sign crops and
machine-readable texts exist, but real line images aligned to their sign sequences are scarce.
We ask: \emph{how well can these resources support sequence recognition, and what does limited
annotation of target tablets add?} Our principal task is to recognize sign sequences from lines
reconstructed by stitching ground-truth sign crops; the output is a sequence of sign names, not
the full scholarly transliteration, which adds phonetic and editorial choices. \rev{Natural line
images (\S\ref{sec:crossperiod}) and complete} tablet-side images with predicted sign locations
\rev{(\S\ref{sec:e2e}) are evaluated separately.}

\paragraph{Constructing compositional supervision.}
We build the missing supervision from real isolated sign crops and text-derived sign
sequences (Figure~\ref{fig:overview}). A text corpus supplies normalized sign sequences; a
tablet-disjoint dictionary supplies real sign crops, plus augmented variants for rare signs.
For each token we sample an exemplar, height-normalize it, and stitch the crops into a line
image labeled with the sequence. The line arrangement is synthetic; its sign crops come from
existing images or their augmentations. We pre-train sequence recognizers on these synthetic pairs (Stage-A)
without paired real-line images, then adapt them on annotated target tablets (Stage-B).
Stage-B adaptation and the principal evaluation also use stitched lines, built from each real
line's own ground-truth crops. Such composition has precedent in other scripts \citep{senthil2020hindi}; here we ask what it
is worth for cuneiform across recognition paradigms, and what the supervision design and the
target annotation each contribute.

\paragraph{Recognition from existing resources and limited annotation.}
We test two connectionist temporal classification (CTC) recognizers, \crnn and \svtr (an SVTRv2-inspired variant), and an autoregressive vision-language model
(VLM), \vlm. On reconstructed lines with a closed vocabulary, adaptation on 100
Ur~III tablets brings all three to 18--25\% sign error rate (SER) in-period. Removing
Stage-A raises SER by \mbox{24--52}\,pp across recognizers and domains
(\S\ref{sec:crossperiod}), including for the VLM with its strong pre-trained base. \rev{For \svtr, the
gain carries over to natural line images cut from the tablet without stitching (in-period SER
52.8\% versus 89.2\% without Stage-A).} Target-tablet adaptation improves in-period SER by 12--15\,pp, but off-domain gains are
smaller and uneven, and cross-collection SER stays at 66--74\% even after
out-of-inventory signs are excluded.

\paragraph{What makes the supervision useful?}
To understand the gain, we compare what the same CTC architecture (\svtr) learns from
individual signs and from composed multi-sign lines. With 100 adaptation tablets, pre-training on shuffled
multi-sign lines lowers SER by 23--46\,pp compared with the same token groups rendered as
single-sign images (Figure~\ref{fig:strict}). The comparison uses test lines that share no exact match or 4-gram
with the source text. Attested co-occurrence and
order add a \rev{further, smaller benefit: positive in-period for every pre-training seed tested, and seed-dependent off-domain.} Given known
sign boundaries, a strong per-sign classifier has lower off-domain SER than \rev{\svtr, but
rescoring the classifier's candidates with \svtr lowers its SER on all three sets}
(\S\ref{sec:crossperiod}).

\paragraph{Contributions.}
(i)~\emph{Compositional supervision for cuneiform}: we turn separately available glyph and
text resources into paired sequence supervision through sign-name mapping, dictionary
construction, rare-sign augmentation, and line rendering (\S\ref{sec:method}).
(ii)~\emph{Effectiveness across recognition paradigms}: in matched ablations, compositional
pre-training lowers SER by 24--52\,pp for \crnn, \svtr, and \vlm across periods and
collections, given limited adaptation on target \rev{tablets; for \svtr it also lowers SER on natural line images} (\S\ref{sec:crossperiod}).
(iii)~\emph{Evidence on supervision and annotation value}: within-\svtr controls show a
large effect of the multi-sign training format (23--46\,pp) and a smaller contribution from
attested text structure; across all three recognizers, adaptation on target tablets helps in-period
recognition far more than off-domain recognition (\S\ref{sec:crossperiod},~\S\ref{sec:strict}).
(iv)~\emph{\rev{Whole-side} recognition}: a modular pipeline with predicted sign
locations and reading order reaches 38.05\% whole-side SER on reused in-period \rev{tablets; its Stage-A
sequence verifier lowers SER on both reused and source-disjoint} tablets (\S\ref{sec:e2e}).

\section{Related Work}
\label{sec:related}

\paragraph{Cuneiform visual recognition.}
Prior work on cuneiform image recognition \citep{bogacz2022survey} spans sign-level tasks: sign
and wedge detection on photographs and 3D renderings
\citep{dencker2020weak,stotzner2023wedge,maicubeda2023,cobanoglu2024signdetection}, isolated-sign
classification \citep{verwimp2025oldbabylonian,mikulinsky2025protosnap}, and segmentation-guided
per-sign classification \citep{poudel2026epigraphnet}.  A second line builds reading-order output
from such predictions: detect-then-classify pipelines such as DeepScribe
\citep{williams2025deepscribe} report a whole-tablet transliteration character error rate, and the
Electronic Babylonian Library pipeline \citep{che2026ebl} scales sign detection to 87k
fragments with heuristic line grouping. It reports normalized $n$-gram set-overlap
of 0.23 on 26,790 sufficiently annotated fragments, excluding training fragments; this
set-overlap measure is not directly comparable to sign-token edit distance. Line-level
transliteration was also attempted from hand-drawn copies with sparse data
\citep{bogacz2017transliteration}. We instead study compositional line-level supervision for
sequence recognizers, evaluated with sign-token edit distance (SER).

\paragraph{Data resources.}
MaiCuBeDa's sign and line annotations on 3D renderings
\citep{homburg2025maicubeda,ahlborn2025maicubedaht} supply real lines for adaptation
and evaluation. The eBL and Old Babylonian sign corpora
\citep{lewenstein2026signdataset,smidt2026obsigns} supply isolated exemplars for composition;
CuneiML \citep{chen2023cuneiml} is a related multimodal tablet collection.

\paragraph{Concurrent tablet-level work.}
\rev{Concurrent work} predicts ordered sign sequences directly from Ur~III tablet-side
images with a single vision-language model trained progressively on signs,
synthetic lines, synthetic tablet sides, and real tablet sides
\citep{anonymous2027spatiallygrounded}.
By contrast, our principal experiments isolate the effect of compositional line
supervision under known sign locations across three recognizers, periods,
and collections. Our whole-side experiment combines separately trained
detection, classification, and sequence-verification components with geometric layout
rules and language-model scoring. The studies share some source resources but test
different training strategies and inference systems. Comparing their SERs requires
aligning evaluation images, sign inventories, reference construction, and scoring conventions.

\paragraph{Synthetic and compositional supervision.}
Synthetic data bootstraps low-resource text recognition
\citep{pippi2025quovadis,baena2024dtlr}; CuReD \citep{gordin2024cured} applies it to
typeset cuneiform transliterations. For tablet images, \citet{rusakov2019gan} generate
signs with adversarial networks; \citet{brandenbusch2021context} generate class-controlled
signs using surrounding visual context. \rev{\citet{bullenkamp2026synthetic} procedurally generate synthetic 3D tablets
for wedge segmentation, sign detection, and sign classification; we instead compose real 2D sign crops into
line-level sequence supervision.}
ProtoSnap \citep{mikulinsky2025protosnap} uses prototype-to-crop skeleton alignments
for structure-conditioned synthesis. StackMix \citep{shonenkov2021stackmix} recombines
fragments weakly segmented from labeled handwriting lines using external text;
\citet{souibgui2022compositional} build lines from generated and augmented symbols.
DTLR \citep{baena2024dtlr} pre-trains on font-rendered and CASIA-based synthetic lines,
then adapts to real line annotations. Closest to our setting,
\citet{senthil2020hindi} compose synthetic Hindi word images from an independent
handwritten-character library and a lexicon to pre-train a CNN--RNN/CTC recognizer for few-instance
adaptation---the same resource-combination logic. What remains unclear for cuneiform is how
much such composition contributes after adaptation to real tablets, and whether line format or
attested sign order produces the benefit (\S\ref{sec:crossperiod}, \S\ref{sec:strict}). For
scene text, \citet{baek2021strfewer} showed that pooled real labeled data trains competitive
recognizers; we measure what target-tablet adaptation adds after compositional pre-training.

\paragraph{VLMs and language priors.}
VLMs support vision-language tasks \citep{liu2023llava,chen2024internvl}, while OCR
benchmarks report difficulties across Unicode scripts and ancient writing
\citep{kargaran2026glotocr,chen2025obibench}. Language priors can affect OCR outputs, with
model-specific visual-grounding failures \citep{karamolegkou2026readguess,liang2026deepseekocr}.
Fine-tuned VLMs and lightweight OCR models are competitive in their settings
\citep{chung2025manchu,ppocr2026}.

\paragraph{Evaluation contamination.}
Our overlap audit connects to work on cross-language recognition, distribution shift, and
evaluation contamination
\citep{crosilla2025benchmarkinghtr,recht2019imagenet,levchenko2025historicalocr}. Synthetic
lines rendered from a text corpus can expose evaluation sign sequences even when tablets and
images are disjoint, so tablet-disjoint splits do not guarantee unseen lines (\S\ref{sec:overlap}).

\section{Task and Available Resources}
\label{sec:task}

\subsection{Recognition Task and Input Conditions}
\label{sec:taskdef}

\emph{Line-level cuneiform sign-sequence recognition} maps a reconstructed tablet line image to
Assyriological sign names (e.g., \texttt{LUGAL}, \texttt{ŠUM\textsubscript{2}},
\texttt{KI}; CDLI/ePSD convention, \citealp{cdli2023}).
The inventory is fixed at 749 labels: 747 canonical sign types, \texttt{X} (illegible), and
\texttt{lacuna} (reconstructed); ``out of vocabulary'' (OOV) means outside this inventory.
Stage-A supervision draws on two independent resources combined in
\S\ref{sec:compositional}---a tablet-disjoint dictionary of real isolated-sign crops and
Ur~III sign sequences from CDLI texts \citep{cdli2023}---while annotated MaiCuBeDa
tablets \citep{homburg2025maicubeda,ahlborn2025maicubedaht} supply Stage-B adaptation and
evaluation (coverage in \S\ref{sec:splits}).

\textbf{Oracle-localized input.} The controlled experiments supply ground-truth (GT) sign
locations. We crop sign boxes derived from GT polygons and order them by horizontal box
center on MaiCuBeDa Hilprecht tablets (in-period and cross-period sets), or by annotation
character index on MaiCuBeDa-HT (cross-collection set). The crops are height-normalized and
stitched into a reconstructed line; OOV signs are excluded from both image and label. These
sign-localized, stitched inputs reflect the current annotation scale and isolate recognition
from localization.

\rev{\textbf{Natural-line input.} Section~\ref{sec:crossperiod} also tests Stage-A on natural line
images: each annotated line is cut from the tablet-side image as one rectangle, without per-sign
cropping or stitching; on Hilprecht lines, signs outside the inventory form an explicit, scored OOV class,
and HT keeps the prefiltered references of its line manifest (Appendix~\ref{app:natural_matched}).}

\textbf{Automatic whole-side input.} The \rev{modular} system (\S\ref{sec:pipeline}) instead reads
tablet-side images with predicted boxes and reading order, without annotation-based filtering
(\S\ref{sec:metrics}, \S\ref{sec:e2e}).

\section{Compositional Supervision}
\label{sec:method}

\subsection{Constructing Synthetic Training Pairs}
\label{sec:compositional}

\begin{figure}[t]
\centering
\includegraphics[width=\linewidth]{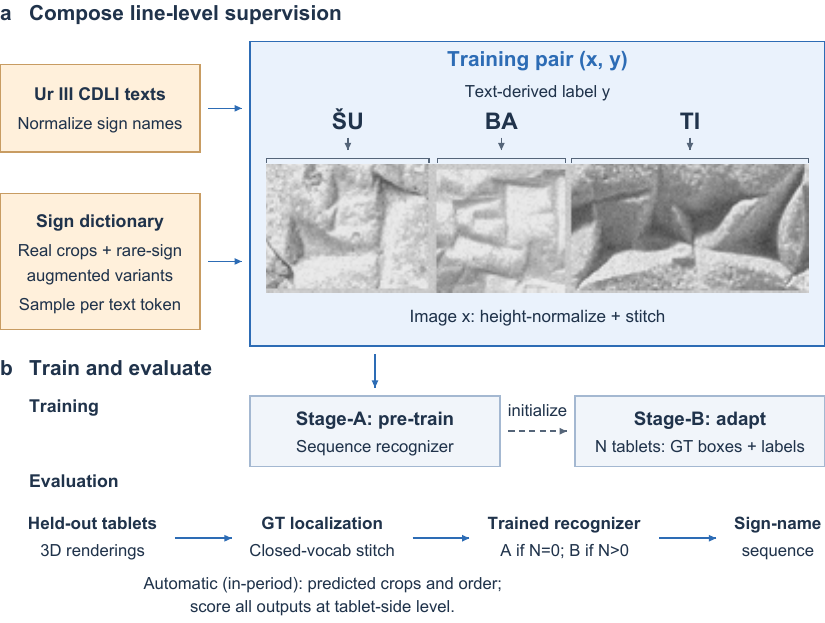}
\caption{\textbf{Synthetic training pairs from real sign crops and existing texts.} \textbf{(a)} Texts supply sign sequences; a dictionary supplies a crop for each token. Height-normalizing and stitching the sampled crops creates a synthetic line image paired with the text-derived label. \textbf{(b)} Stage-A pre-trains the recognizer; Stage-B adapts on $N$ annotated Ur~III tablets ($N{=}0$ skips only Stage-B). Primary evaluation uses GT-localized, closed-vocabulary stitches of 3D renderings; the \rev{modular} pipeline with predicted localization is in \S\ref{sec:pipeline} and Figure~\ref{fig:auto_verifier} (Appendix~\ref{app:e2edetail}).}
\label{fig:overview}
\end{figure}

\noindent
We synthesize labeled line images for Stage-A pre-training as follows
(Figure~\ref{fig:overview}a).  A sequence of canonical sign names is taken from an Ur~III
CDLI text and mapped to the label inventory (\S\ref{sec:taskdef}).  For each sign, we sample one crop uniformly from its dictionary entry.
The dictionary contains real glyph crops; under-represented sign types also include geometric
and photometric augmentation variants within existing classes (Appendix~\ref{app:ht}).
We height-normalize each crop to 64 pixels and concatenate the crops
left-to-right into a \emph{stitched} line image.
The normalized sign-name sequence from the text supplies its label; OOV signs are excluded
from both image and label. The dictionary
holds 199{,}243 images covering 744 of the 747 canonical sign types and contains no
MaiCuBeDa-tablet images (provenance in Appendix~\ref{app:ht}). The three empty entries
(\texttt{du}, \texttt{šar\textsubscript{2}}, \texttt{600}) occur in no Stage-A line or
evaluation reference: normalization maps \texttt{du} and \texttt{šar\textsubscript{2}} to
\texttt{DU} and \texttt{ŠAR\textsubscript{2}}, which have exemplars, and drops bare numerals.

\textbf{Synthetic corpus and evaluation separation.} After removing \rev{the composed lines} whose
CDLI text \rev{our tablet-to-text mapping links} to a development, in-period, or cross-period tablet (Haft
Tappeh texts are already \rev{absent; Table~\ref{tab:overlap} lists two texts the mapping missed),} the Stage-A corpus contains 387{,}315 composed image--sequence pairs
(deletion counts in Appendix~\ref{app:strict}). They average 4.6 signs and retain all 285
corpus sign types. ``Text-disjoint'' means no shared CDLI \rev{\emph{text} (apart from these two):} repeated formulae can
still reproduce an evaluation line verbatim (\S\ref{sec:overlap}). All primary results use
Stage-A models trained on this corpus. CTC models split it into 347{,}322 training and
39{,}993 validation pairs; \vlm uses 126{,}000 training records with low-rank adaptation (LoRA), followed by the
continuation documented in Appendix~\ref{app:training}. Texts of the Stage-B training-pool
tablets are not removed.

\subsection{Sequence Recognizers and Target-Tablet Adaptation}
\label{sec:models}

The primary experiments use three sequence recognizers: two CTC models---\crnn
\citep{shi2016endtoend} and our \svtr---and the autoregressive \vlm.

\textbf{CTC recognizers.}  \crnn and \svtr map a stitched line image to the sign sequence
by CTC decoding (749 labels + blank); no sign-boundary labels are given to the decoder.  The
asterisk in \svtr marks our SVTRv2-\emph{inspired} variant, not the original: SVTR mixing
blocks and SVTRv2 vertical feature rearrangement, but \emph{no} semantic-guidance module
\citep{du2022svtr,du2024svtrv2}.

\textbf{Vision-language model.}  \vlm \citep{paddleocrvl2026} is a 0.9\,B multimodal
model with a visual encoder and autoregressive decoder; Stage-A and Stage-B fine-tune
LoRA adapters \citep{hu2022lora}; configuration and decoding are in
Appendix~\ref{app:training}. Its smaller Stage-A budget (\S\ref{sec:compositional}) is a
confound for cross-model comparison.

\label{sec:stageB}
Stage-B fine-tunes each recognizer on $N$ real MaiCuBeDa tablets.
Inputs are GT-localized sign crops from each annotated line, stitched from left to right by bounding-box center.
CTC updates all parameters; \vlm continues LoRA from the Stage-A adapter.
Details are in Appendix~\ref{app:recognizer_recipe}.

\subsection{Recognition with Predicted Localization}
\label{sec:pipeline}

For whole-side evaluation, we assemble separately trained components. Given a tablet-side image and its surface type,
a YOLO26n detector \citep{jocher2026yolo26} proposes sign boxes, a dictionary-pre-trained DeiT3
classifier \citep{touvron2022deit3} proposes candidate labels, geometric layout rules group and
order the boxes into lines, and a base decoder selects label sequences using classifier and
six-gram language-model (LM) scores with weights selected on the development (DEV) set. A compositionally pre-trained $N{=}100$ \svtr then
verifies candidate labels on their own crops, adding full CTC probability with a DEV-selected
weight (Figure~\ref{fig:auto_verifier} in Appendix~\ref{app:e2edetail}). Inference uses no reference boxes, lines, or text; the
classifier uses no synthetic-line pre-training.
This modular inference system differs from the direct tablet-side image-to-sequence model
studied in concurrent work \citep{anonymous2027spatiallygrounded}.

\newcommand{\gtc}[1]{\textcolor{gray}{#1}}
\newcommand{\okc}[1]{\textcolor{fgreen}{#1}}
\newcommand{\erc}[1]{\textcolor{fred}{\underline{#1}}}
\newcommand{\gap}{\textcolor{gray}{\textendash}}

\section{Evaluation Protocol}
\label{sec:setup}

\subsection{Evaluation Sets and Annotation Budgets}
\label{sec:splits}

\textbf{Training and selection (Ur~III, MaiCuBeDa Hilprecht).}
For line recognition, Stage-B samples $N$ tablets from a pool of 214 Ur~III tablets. A
tablet-disjoint 50-tablet development (DEV) set (309 CTC lines, 342 VLM lines) selects
learning rates and early stopping. Because the VLM's mapping also scores numerals, its in-period, cross-period, and
HT line counts are 718, 1{,}254, and 1{,}408 versus 670, 1{,}243, and 1{,}320 for CTC
(Appendix~\ref{app:ht}). Primary line evaluation excludes lines with fewer
than three in-vocabulary signs; whole-side evaluation keeps all annotated in-inventory signs.

\textbf{Automatic-system budgets and selection.} The detector trains on 450 tablets and
validates on 50, DeiT3 adapts on the 214-tablet pool, and the recognizers on a 100-tablet
subset. Of the 50 DEV tablets, 43 are in detector training and 7 in detector validation, while
DEV also selects the system's checkpoints and decoding weights. These selections are
therefore conditional on a detector that has seen DEV; the \rev{in-period and source-disjoint
evaluation tablets (\S\ref{sec:e2e})} are outside
detector training, detector validation, and DEV (Appendix~\ref{app:auto_refinement}).

\textbf{Three primary evaluation sets} are disjoint from recognizer training and DEV at the
tablet level \rev{and, apart from one cross-period text,} from Stage-A at the CDLI-text level (\S\ref{sec:overlap}). Sets (2) and (3)
were not used to tune or select the primary recognizers; the generalization claims rest on them.

(1)~\emph{In-period (reused; Ur~III)}: 100 tablets (198 tablet-side images, ``stems''),
670 lines, same period and collection as training; these lines featured in earlier in-period
analyses and were inspected during development, so we label them
\emph{reused} rather than pristine.

(2)~\emph{Cross-period}: 192 MaiCuBeDa tablets (1{,}243 \rev{lines): 191 from} six non-Ur~III periods, up to $\sim$500 years from training \rev{(Appendix~\ref{app:perperiod}),
and one Ur~III tablet, HS~1032, that our tablet-to-text mapping missed (Table~\ref{tab:overlap}).} The set was
assembled without inspecting model outputs.
Cross-period transfer is relative to the Ur~III adaptation period: the crop dictionary spans
multiple periods, with most crop-period labels unknown, so Stage-A is not period-disjoint.

(3)~\emph{Cross-collection}: MaiCuBeDa-HT \citep{ahlborn2025maicubedaht}, the
Haft Tappeh collection of Middle-Elamite-period Akkadian tablets (c.\ 1500--1400\,BC),
annotated by a different campaign.  The CTC evaluator covers 1{,}320 lines on 305 tablets. Of the annotated tokens, 18.5\% are outside our inventory and are
excluded from image and label, as in the other sets under the oracle protocol.  HT differs
from training in site, language, period, scribal tradition, and annotation
campaign (a different imaging batch under the same \rev{Multi-Scale Integral Invariant (MSII)} rendering technology).

\begin{samepage}
Frozen manifests fix all three sets (cross-period and HT hashes in Appendix~\ref{app:training}). The 670-line in-period set instantiates only 147 of
the 749 labels (93 with $\geq$5 occurrences), the cross-period set 212, and the
cross-collection set 115; performance covers these instantiated subsets, not the full
inventory. The 285 types in Stage-A text cover
100/99.95/96.7\% of in-period/cross-period/HT evaluation tokens.
\par\end{samepage}

\subsection{Comparisons and Model Selection}

\paragraph{Primary protocol (matched-protocol ablation).}
For each recognizer and annotation budget $N\in\{25,100\}$, we train one arm with Stage-A
pre-training and one without it. Both arms use the same Stage-B tablets, seeds
$\{42,123,456\}$, batch size, epoch budget, and patience (CTC: 100 epochs/15; \vlm: 40/10).
Only the learning rate may differ: each arm selects it by seed-42 DEV \lcs from a fixed grid,
then applies it to all seeds. Appendix~\ref{app:training} gives the grids and the VLM's reuse
of previously DEV-selected rates after text-disjoint retraining. Without Stage-A, CTC models
start randomly and the base VLM receives a fresh LoRA adapter.
Analysis and cross-period evaluation plans were locked
before primary evaluation (Reproducibility Statement).

\subsection{Text-Overlap Controls}
\label{sec:overlap}

All primary results (\S\ref{sec:crossperiod}), strict subsets, and automatic sequence-model
results use \emph{text-disjoint} Stage-A models (\S\ref{sec:compositional}).
\textbf{Text-level deduplication does not remove recurring formulae.}
Stage-A contains \rev{no line} of the CDLI texts linked to DEV, in-period, or cross-period \rev{tablets, except
two texts our mapping missed (cross-period HS~1032, 15 Stage-A lines; DEV HS~2445, 4 lines;
Table~\ref{tab:overlap}).} It also contains none of the 282 HT texts,
because it draws only on Ur~III texts. Yet 48.6\% of in-period lines still recur verbatim because administrative formulae
appear across tens of thousands of texts. Cross-period and HT lines recur far less
(Table~\ref{tab:overlap}).

We therefore recompute every contrast on a \emph{strict} subset requiring \emph{both}
conditions: a line shares \emph{no 4-gram} with any Stage-A line \emph{and} its exact
sequence never occurs in the Stage-A corpus (Table~\ref{tab:strictser},
Appendix~\ref{app:strict}).  Both are needed because a short line ($<$4 signs, 45\% of the
in-period set) has no 4-gram and would pass a 4-gram filter even when it recurs verbatim;
the exact-unseen condition removes these.  The in-period strict subset is
43\% of lines ($n$=290 in the annotation audit); a second computation on the SER
records uses raw sign names and reports its own subset sizes (Appendix~\ref{app:strict}).
\begin{table}[!b]
\centering
\footnotesize
\setlength{\tabcolsep}{4pt}
\caption{\textbf{Synthetic pre-training benefit at $N{=}100$.}
SER$\downarrow$ (\%) after Stage-B adaptation on 100 tablets with and without Stage-A
pre-training on the synthetic pairs (\S\ref{sec:compositional}); $N{=}0$ is Stage-A alone.
GT-localized, closed-vocabulary inputs; means over seeds 42/123/456. $\Delta$ = without $-$
with Stage-A, with paired-tablet 95\% CIs conditional on fixed runs. Line counts differ by
model (\S\ref{sec:splits}), so cross-model values are \rev{descriptive; a common subset with identical
references gives removal costs within 1.3\,pp of these (Appendix~\ref{app:commonsubset}).} $N{=}25$:
Table~\ref{tab:smallbudget}; per-seed values: Appendix~\ref{app:perseed}.}
\label{tab:main}
\par\vspace{\baselineskip}
\begin{tabular*}{\linewidth}{@{\extracolsep{\fill}}llcccc@{}}
\toprule

Set & Model & $N$=0 & with St.-A & no St.-A & $\Delta$ (pp) [95\% CI] \\
\midrule
\multirow{3}{*}{\shortstack[l]{In-period\\(Ur~III)}}
& \svtr & \Tnum{2.1}{33.6} & \Tnum{2.1}{21.5} & \Tnum{2.1}{72.5} & \Tci{+2.1}{+50.9}{48.5}{53.3} \\
& \crnn & \Tnum{2.1}{37.5} & \Tnum{2.1}{24.8} & \Tnum{2.1}{73.1} & \Tci{+2.1}{+48.3}{45.9}{50.6} \\
& \vlm  & \Tnum{2.1}{32.8} & \Tnum{2.1}{17.9} & \Tnum{2.1}{45.8} & \Tci{+2.1}{+27.9}{26.0}{29.9} \\
\midrule
\multirow{3}{*}{\shortstack[l]{Cross-\\period}}
& \svtr & \Tnum{2.1}{40.7} & \Tnum{2.1}{38.4} & \Tnum{2.1}{90.8} & \Tci{+2.1}{+52.4}{49.3}{54.9} \\
& \crnn & \Tnum{2.1}{48.3} & \Tnum{2.1}{41.9} & \Tnum{2.1}{90.5} & \Tci{+2.1}{+48.6}{45.7}{50.9} \\
& \vlm  & \Tnum{2.1}{41.1} & \Tnum{2.1}{41.2} & \Tnum{2.1}{76.2} & \Tci{+2.1}{+35.0}{32.5}{37.0} \\
\midrule
\multirow{3}{*}{\shortstack[l]{Cross-\\collection}}
& \svtr & \Tnum{2.1}{70.4} & \Tnum{2.1}{71.2} & \Tnum{2.1}{97.8} & \Tci{+2.1}{+26.7}{24.9}{28.6} \\
& \crnn & \Tnum{2.1}{72.5} & \Tnum{2.1}{73.6} & \Tnum{2.1}{97.5} & \Tci{+2.1}{+23.9}{22.1}{25.6} \\
& \vlm  & \Tnum{2.1}{69.6} & \Tnum{2.1}{66.1} & \Tnum{2.1}{90.1} & \Tci{+2.1}{+23.9}{22.3}{25.6} \\
\bottomrule
\end{tabular*}

\end{table}

\subsection{Metrics and Uncertainty}
\label{sec:metrics}

\paragraph{Primary and secondary metrics.}
Our primary metric is \textbf{Sign Error Rate} (SER$\downarrow$), the corpus-level
sign-token Levenshtein distance
\begin{equation*}
\mathrm{SER}=\frac{\sum_i \left(S_i+D_i+I_i\right)}{\sum_i |y_i|},
\end{equation*}
where $S_i$, $D_i$, and $I_i$ count substitutions, deletions, and insertions against
reference $y_i$; unlike LCS Recall, SER penalizes insertions.
\emph{Oracle-line SER} scores saved outputs on the reconstructed lines; \vlm
outputs are normalized and vocabulary-filtered beforehand.
\emph{Automatic whole-side SER} concatenates all output rows of a side without spatial
masking, retaining unknown and unmatched output tokens. Its fixed reference has 3,280
in-inventory signs; 903 annotated regions whose readings map outside the inventory (e.g.,
numeral notations) are excluded from the reference only, so outputs produced there can still
incur errors. It therefore measures a closed-inventory target, not full transcription, and is not
comparable to oracle-line SER (Appendix~\ref{app:auto_scoring}). \textbf{Exact-Line Accuracy}, LCS Recall, and length
diagnostics are defined in Appendix~\ref{app:strict}.

\paragraph{Uncertainty and multiple comparisons.}
Primary ablation contrasts compare 3-seed means using paired tablet-level bootstrap
($B$=10{,}000, percentile 95\% CI), resampling the same tablets across fixed runs.
The intervals condition on fixed training runs: seeds are not resampled, and Stage-A
retraining variance is \rev{excluded; \S\ref{sec:strict} adds an across-seed interval for the attested-order
contrast.} The 95\% CIs are unadjusted for multiple comparisons.
All 18 contrasts in Tables~\ref{tab:main} and~\ref{tab:smallbudget} have CIs excluding zero.

\section{Results}
\begin{figure}[t]
\centering
\includegraphics[width=\linewidth]{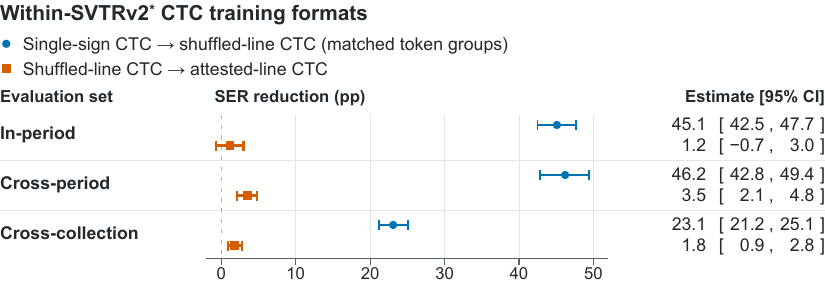}
\caption{\textbf{Training-format contrasts} within \svtr ($N{=}100$; 3-seed Stage-B mean). Positive favors the second format; bars: paired-tablet 95\% CIs given \rev{the seed-42 Stage-A checkpoints (further seeds: Table~\ref{tab:stageaseeds}).} Values are on strict-subset lines (exact-unseen, no shared 4-gram). Canvas, target length, and alignment co-vary in the first contrast; co-occurrence and order in the second. They do not causally decompose Table~\ref{tab:main} (Table~\ref{tab:strictci}).}
\label{fig:strict}
\end{figure}
\label{sec:results}

\subsection{Recognition, Adaptation, and Transfer}
\label{sec:crossperiod}

\paragraph{Absolute recognition.}
With oracle sign localization and closed-vocabulary, stitched inputs, adaptation on 100 real
Ur~III tablets brings the models to 17.9--24.8\% SER in-period, 38.4--41.9\% cross-period, and
66.1--73.6\% cross-collection (Table~\ref{tab:main}; qualitative examples in
Figures~\ref{fig:qualitative} and~\ref{fig:qualitative_diagnostics}). Exact-line accuracy falls from 42--57\% in-period to 1--3\%
cross-collection. The reported HT SER applies only to in-inventory signs. Another 18.5\% of
annotated HT tokens are excluded from its reference and input, so the result does not measure
recognition of the full collection.

\paragraph{Value of synthetic pre-training.}
Removing Stage-A raises SER by 24--52\,pp at $N$=100 (Table~\ref{tab:main}). The gaps are smallest for \vlm and on HT, where absolute SER
remains high for every model. For \vlm, whose without-Stage-A arm keeps the strong pre-trained base
with a fresh LoRA adapter, removing Stage-A still costs 24--35\,pp. At $N$=25 the gaps are larger (up to 68\,pp; Table~\ref{tab:smallbudget}); there
the no-Stage-A \crnn collapses to near-total error (SER ${\approx}$ 100--102\%),
and completing the full 100-epoch schedule still leaves it weak
(Appendix~\ref{app:additional_checks}).

\paragraph{The gain survives a converged control and a strict text subset.}
For \svtr, the Stage-A advantage remains large against a 300-epoch from-scratch baseline:
$+$46.7/$+$50.7/$+$25.2\,pp at $N{=}100$ (Appendix~\ref{app:perseed}).
On strict text subsets, the removal costs stay within ${\sim}2$\,pp of the all-line costs
across all nine $N$=100 contrasts (Table~\ref{tab:strictser}).

\paragraph{\rev{The gain transfers to natural line images.}}
\rev{In the matched \svtr ablation on natural lines (\S\ref{sec:taskdef}, Appendix~\ref{app:natural_matched}),
Stage-A lowers SER from 89.2\% to 52.8\% on all 906 annotated in-period lines (95\% CI of the difference
$[34.2,38.7]$; 36.2\,pp on the 670 primary lines) and by 19.0 and 7.0\,pp cross-period and cross-collection; a 300-epoch no-Stage-A
control is still 32.4\,pp worse in-period. The gain holds without line annotation at test time: with
line regions from the automatic system's predicted rows or from an independently trained line
segmenter, Stage-A lowers in-period whole-side SER by 34.9 and 29.5\,pp (outputs tagged OOV excluded;
Table~\ref{tab:natural_auto}).}

\paragraph{Adaptation and transfer.}
Without target tablets ($N$=0), Stage-A models reach 33--72\% SER. Adapting on 100 Ur~III tablets cuts in-period SER by
12--15\,pp, to 18--25\%. Off-domain reductions are smaller:
$-$0.2 to $+$6.3\,pp cross-period and $-$1.1 to $+$3.5\,pp cross-collection.
Further Ur~III
supervision mostly buys in-distribution refinement, while Stage-A already supplies most off-domain
capability at $N$=0 (the composed corpus covers nearly all evaluation tokens; for the VLM, partly
from its pre-trained base).  Cross-period performance is heterogeneous across the six periods
(Appendix~\ref{app:perperiod}). Cross-model differences are descriptive:
initialization, capacity, token mapping, and Stage-A budget differ (Appendix~\ref{app:perseed}).

\paragraph{Strong per-sign classifier baseline.}
A dictionary-trained DeiT3-medium classifier \citep{touvron2022deit3}, adapted on the same
100 tablets, classifies each GT sign crop in line order. Evaluation lines, labels,
OOV rule, and strict subsets match \svtr exactly. Given true sign boundaries, DeiT3
performs better than \svtr on the two off-domain sets (Table~\ref{tab:persign_main}); on
strict in-period lines its SER is 22.2\% versus 25.4\% ($\Delta=-3.2$\,pp, 95\% CI
$[-5.5,-0.8]$). Different architectures, initializations, and objectives make this a pipeline
comparison, not an isolated test of sequence context (Appendix~\ref{app:persign}). \rev{The two models
are complementary: rescoring DeiT3's candidate sequences with the Stage-A \svtr's CTC probability
lowers SER below DeiT3 alone on all three sets (to 14.6\% in-period, and from 22.2\% to 17.0\% on strict
in-period lines), while rescoring with the
no-Stage-A \svtr gives no off-domain gain (Table~\ref{tab:persign_main},
Appendix~\ref{app:persign_fusion}). \svtr also needs about 28$\times$ fewer FLOPs per line and runs
about 12$\times$ faster on CPU (Appendix~\ref{app:speed}). With predicted boundaries, it serves as a
verifier (\S\ref{sec:e2e}).}

\begin{table}[t]
\centering\small
\setlength{\tabcolsep}{5pt}
\caption{\textbf{Strong per-sign comparison under the same oracle protocol.}
$N{=}100$, three-seed SER (\%); $\Delta$ = per-sign $-$ \svtr; paired-tablet 95\% CIs
condition on fixed checkpoints. \rev{DeiT3+\svtr: DeiT3 candidates rescored by the Stage-A \svtr
(Appendix~\ref{app:persign_fusion}).} Strict subsets and exact-line scores: Table~\ref{tab:persign}.}
\label{tab:persign_main}
\par\vspace{\baselineskip}
\begin{tabular*}{\linewidth}{@{\extracolsep{\fill}}lcccc@{}}
\toprule
Set & Per-sign DeiT3 & \svtr & $\Delta$ (pp) [95\% CI] & \rev{DeiT3+\svtr}\\
\midrule
In-period & \Tnum{2.1}{21.0} & \Tnum{2.1}{21.5} & \Tci{+2.1}{-0.5}{-1.8}{0.9} & \rev{\textbf{14.6}} \\
Cross-period & \Tnum{2.1}{27.8} & \Tnum{2.1}{38.4} & \Tci{+2.1}{-10.6}{-12.2}{-8.9} & \rev{\textbf{25.1}} \\
Cross-collection & \Tnum{2.1}{62.4} & \Tnum{2.1}{71.2} & \Tci{+2.1}{-8.8}{-9.9}{-7.7} & \rev{\textbf{61.7}} \\
\bottomrule
\end{tabular*}
\end{table}

\subsection{Contributions of Line Format and Attested Text Structure}
\label{sec:strict}

\paragraph{Which parts of the synthetic-data construction help?}
We train the same CTC architecture (\svtr) on single signs, on shuffled multi-sign lines, and
on lines following attested text. These Stage-A models form a separate checkpoint family (the attribution family): the
arms share the renderer and match token exposure and optimizer updates, but not compute, and
then follow the locked Stage-B protocol. Shuffling permutes tokens within each split and
re-cuts lines, changing co-occurrence as well as order. Construction, per-arm results,
and all-lines contrasts are in Appendix~\ref{app:shuf}.

\paragraph{Training-format effects within \svtr.}
The largest tested step is from single signs to multi-sign lines. A parent-matched control renders
the same shuffled token groups as separate one-sign images (``singleton-matched'').
At $N{=}100$, shuffled-line pre-training lowers strict-subset SER
by 45.1, 46.2, and 23.1\,pp in-period, cross-period, and cross-collection, respectively
(Figure~\ref{fig:strict}). \rev{Grouping alone therefore does not explain the gain. Canvas, target
length, and CTC alignment co-vary in this contrast, and the singleton-matched model under-emits on
multi-sign lines; but on lines where both models output exactly the reference length, per-position
accuracy is still 86.7\% versus 63.8\% in-period, so output length} alone does not explain the \rev{gap
either (Appendix~\ref{app:length}).} This compares CTC pre-training formats, not per-sign
classification (\S\ref{sec:crossperiod}).

\paragraph{\rev{Attested text structure adds a smaller benefit, consistent in-period.}}
Compared with shuffled re-composition, attested co-occurrence and order \rev{lower SER further. We
trained both Stage-A models with two more seeds (Appendix~\ref{app:stageaseeds}; seed 42 is the run
in Figure~\ref{fig:strict} and Table~\ref{tab:strictci}). In-period, the mean reduction is 4.6\,pp on
all lines, with every per-seed 95\% CI above zero (across-seed 95\% interval $[3.1,6.2]$), and
2.2\,pp on the strict subset, positive for every seed. Off-domain the effect is seed-dependent: the
mean strict-subset reductions are 1.3\,pp cross-period and 1.0\,pp cross-collection, and seed 456
reverses the gain in both sets.}

\subsection{Modular End-to-End Recognition}
\label{sec:e2e}

\rev{On whole sides,} the full modular system reaches \textbf{38.05\%} SER,
versus \textbf{68.92\%} for the strongest original geometric route (Table~\ref{tab:auto-whole-side}),
a \textbf{30.87\,pp} difference (mean of three runs that vary the classifier and verifier seeds;
paired 95\% CI $[28.10,33.88]$).
This comparison evaluates complete systems; it does not isolate the effect of any one module.
The original baselines apply compositionally pre-trained $N{=}100$ recognizers to YOLO-grouped
crops. They share its detector weights, but proposals, crop selection, ordering, decoding, and component budgets differ. All systems are
scored on the same 198 sides of the 100 reused in-period tablets; the reference excludes 903 out-of-inventory
regions, but every output is scored (\S\ref{sec:metrics}). DEV-based selections are
conditional on detector exposure (\S\ref{sec:splits}).

\input{tables/automatic_comparison.tex}

\paragraph{\rev{What Stage-A adds to the system.}}
\rev{The classifier has no synthetic-line pre-training, so Stage-A enters the full system only through
the \svtr verifier; the rest of the 30.87\,pp difference comes from the other changes between the two
systems, which this comparison does not separate. Holding them
fixed within each run (detector, classifier, candidate pool, layout, LM, and decoder weights),
removing the verifier raises mean SER from 38.05\% to \rev{39.88\% (error types:
Table~\ref{tab:verifier_extension}; per-pair results: Appendix~\ref{app:e2edetail}). Most of
this gain requires Stage-A. When the verifier weight is re-selected by the same DEV rule for both models, the matched
no-Stage-A \svtr of Table~\ref{tab:main} receives weight 0 in two of three runs, and the Stage-A
verifier gives 1.48\,pp lower SER (95\% CI $[0.95,2.05]$); at the frozen weight for both, 1.28\,pp
($[0.66,1.92]$). As an independent decoder} on the base
decoder's selected crops and \rev{order, \svtr scores 46.39\% with Stage-A and 82.22\% without it, and
43.88\% versus 76.54\% with the six-gram LM
(Table~\ref{tab:auto-shared-controls}, Appendix~\ref{app:sourcedisjoint}).}}

\paragraph{\rev{Source-disjoint tablets.}}
\rev{We also test 34 Ur~III sides (18 tablets and envelopes, 13 texts) from HeiCuBeDa
\citep{mara2019heicubeda} whose images enter no training, development, or other test data and no
dictionary source. References come from CDLI transliterations, which list more signs than box
annotations (on in-period sides they raise the same system's SER by 5.8\,pp), and the LM is rebuilt
without these texts; Stage-A is not text-disjoint from them (100 of its lines share these texts, rendered
from other tablets' crops; Appendix~\ref{app:sourcedisjoint}). The Stage-A verifier lowers the base
system's SER from 48.66\% to 46.49\% (95\% CI of the difference $[0.64,3.66]$, CDLI-text bootstrap)
and, under the same DEV rule, beats the no-Stage-A verifier by 1.29\,pp ($[0.13,2.32]$). Reading the
system's predicted rows, the natural-line \svtr of \S\ref{sec:crossperiod} reaches 58.2\% whole-side
SER on these sides with Stage-A and 92.6\% without it (outputs tagged OOV excluded;
Table~\ref{tab:natural_auto}).}

\section{Conclusion and Limitations}
\label{sec:conclusion}

Composing real sign crops according to text-corpus sequences supplies pre-training
supervision without paired real-line images: under oracle localization, removing it costs
24--52\,pp SER across three recognizers and three \rev{domains, and for \svtr the gain carries over to
natural line images.} Within \svtr, the multi-sign line
format matters more than attested sign order and co-occurrence. Target-tablet annotation mainly
helps in-period, and cross-period and cross-collection recognition remain difficult. Given true
sign boundaries, a per-sign classifier is better off-domain than \rev{\svtr, but rescoring its candidates
with the Stage-A \svtr beats the classifier alone on all three sets.} With predicted localization, the
\rev{modular} system reduces whole-side SER from 68.92\% to 38.05\% versus the best original geometric
\rev{baseline; its sequence verifier helps more with Stage-A than without it and also lowers SER on source-disjoint tablets.} For Assyriology, composed supervision could help produce
searchable sign sequences for digitized tablets and align them with existing editions,
subject to expert checking.

\textbf{Limitations.}\label{sec:limitations}
Primary evaluation uses GT-localized, fixed-vocabulary stitches of MaiCuBeDa
renderings; collection shifts vary site, language, and campaign, not imaging domain.
The multi-period dictionary may expose styles of unseen tablets/texts
(\S\ref{sec:splits}). Automatic evidence \rev{comes mainly from reused in-period tablets, with a 34-side
source-disjoint set scored against transliterations; component selection is conditional on detector
exposure, and the fixed-inventory reference excludes} some annotated regions. Recognition of out-of-inventory signs and numeral notations, and evaluation
under new imaging conditions, remain open.

\section*{AI Use Statement}
Generative AI tools were used to assist with code implementation, experiment
orchestration, refinement of experimental designs, analysis and verification of
results, literature search, and drafting of exposition.  All AI-assisted code and
analyses were reviewed and tested by the authors, every quantitative result reported
here was produced by the authors' own experimental pipeline, and all research
questions, claims, and conclusions are the authors' own.  We take responsibility
for the final content of this work.

\section*{Ethics Statement}
We use the publicly released MaiCuBeDa Hilprecht \citep{mara2023maicubedadata,
homburg2025maicubeda}\rev{,} MaiCuBeDa-HT \citep{ahlborn2025maicubedaht}\rev{, and HeiCuBeDa Hilprecht
\citep{mara2019heicubeda}} corpora (CC BY-SA 4.0) and CDLI texts \citep{cdli2023}; isolated-sign exemplars come from
the eBL annotated sign-image dataset \citep{lewenstein2026signdataset}
(CC BY-NC 4.0) and an Old Babylonian sign dataset \citep{smidt2026obsigns}
(CC BY 4.0).  No human subjects were involved; the
work aims to accelerate scholarly access to cuneiform sources, and outputs are
sign-name sequences that require expert review before use.

\section*{Reproducibility Statement}
\textbf{Locked plan.}  The primary-analysis protocol was fixed before any
primary-protocol evaluation: a written analysis plan (2026-08-08) set the LR
grids, epoch budgets, patience, seeds $\{42,123,456\}$, the DEV-only
LR-selection rule, and \rev{a Holm family (the intervals reported here are unadjusted; \S\ref{sec:metrics}),} and a locked evaluation plan
(2026-08-13) recorded the cross-period manifest.  These are internally
generated artifacts whose hashes identify the exact versions used
(Appendix~\ref{app:training}), not an externally time-stamped registration.
\textbf{Text-disjoint retraining.}  After the audit of \S\ref{sec:overlap} the
Stage-A corpus was deduplicated, all three Stage-A models were retrained, and
the Stage-B runs were repeated as described in \S\ref{sec:setup}.  Training-run
details, compute, artifact hashes, deviations from bitwise replay, and the HT
manifest construction are in Appendices~\ref{app:training} and~\ref{app:ht}.
\textbf{Code and materials.} \rev{We will release a code package with runnable entry points for
dictionary construction and line synthesis (including the attribution variants), Stage-A/B training,
inference, and scoring, and for every experiment in Appendices~\ref{app:natural_matched}
and~\ref{app:stageaseeds}--\ref{app:length}, with configurations,
seeds, manifests, label maps, a CPU smoke test, and saved per-line records from which all main-text tables and most appendix tables can be recomputed without retraining. It will
not include raw images, model weights, or the external line segmenter's weights and outputs; source data are available
from the cited repositories under their licenses.} Six of the 108 main-ablation run$\times$set records---the
without-Stage-A \crnn $N$=25 runs for seeds 123/456 on three sets---remain explicit failure entries scored by
the empty-output convention; no model output is synthesized for them. Missing historical run metadata,
including the per-sign baseline's gaps (Appendix~\ref{app:persign}), remains identified as \rev{missing.}
Third-party material retains its source attribution and applicable terms.

\bibliographystyle{iclr2027_conference}
\bibliography{references}

\appendix
\raggedbottom
\makeatletter
\setlength{\@fptop}{0pt}
\setlength{\@fpsep}{12pt plus 4pt}
\setlength{\@fpbot}{0pt plus 1fil}
\makeatother

\noindent\textbf{Appendix guide.} \rev{Appendices~\ref{app:strict}--\ref{app:training}} cover overlap controls and
strict subsets \rev{(Table~\ref{tab:strictser}), data provenance, and} training and compute \rev{(Table~\ref{tab:trainconfig});}
Appendix~\ref{app:e2edetail} gives the automatic pipeline, component budgets, and whole-side \rev{results;
Appendices~\ref{app:perseed} and~\ref{app:perperiod} give per-seed results and period breakdowns.}
Appendix~\ref{app:shuf} gives the primary attribution study followed by a separate, stronger global-shuffle
control. The VLM rank and image-sensitivity probes \rev{(Appendix~\ref{app:vlm})} use earlier text-exposed checkpoints
and their stated auxiliary metrics. \rev{Appendix~\ref{app:natural_matched} gives the matched natural-line comparison,
Appendix~\ref{app:persign} the per-sign baseline, and Appendix~\ref{app:additional_checks} the} CRNN
stopping-condition and VLM DEV \rev{checks. Appendices~\ref{app:stageaseeds}--\ref{app:length} give independent Stage-A
seeds, the source-disjoint whole-side evaluation, the per-sign/sequence combination, inference cost, a common
CTC/VLM subset, and a length/alignment analysis.}

\section{Overlap Audit and Overlap-Controlled Subsets: Full Tables}
\label{app:strict}

\paragraph{Auxiliary metrics (definitions).}
\textbf{Exact-Line Accuracy} is the fraction of lines with $S_i+D_i+I_i=0$, i.e., whose
predicted sign sequence equals the reference.
\textbf{LCS Recall} measures ordered sign recovery:
$\sum_i \text{LCS}(\hat{y}_i^{V},y_i)/\sum_i|y_i|$.
Here LCS is the longest common subsequence and $\hat{y}_i^{V}$ is the prediction
projected to the 749-label vocabulary. Counts are pooled across lines.
This metric gives partial credit but does not penalize insertions.
For the $N$=100 with-Stage-A runs in Table~\ref{tab:main}, we compute
$\sum_i|\hat{y}_i|/\sum_i|y_i|$ per seed and average over three seeds.
Across the nine recognizer$\times$set cells, this ratio ranges from 0.974
(\crnn, cross-collection) to 1.001 (\svtr, cross-collection).
There is no systematic over-generation, but insertions and deletions can offset
each other in the length ratio. SER therefore remains the primary metric.

\begin{table}[H]
\centering
\footnotesize
\setlength{\tabcolsep}{5pt}
\caption{\textbf{Overlap between the text-disjoint Stage-A corpus and each evaluation set.}
The last column is the \emph{strict}-subset fraction (no shared 4-gram \emph{and} no
verbatim match); the no-4-gram-only fractions are higher (in-period 72\%).  In-period
percentages are from the annotation-audit script (671 lines); the SER-record
pipeline counts 670 lines and reports its own subset sizes separately (\S\ref{app:strict}).
(Before deduplication: 99/100 in-period and 47/50 development
texts were in Stage-A, with 56.2\% and 42.4\% verbatim lines.  Deduplication removed 859
lines, 0.22\%, from 146 texts of the 388{,}174 composed lines, leaving 387{,}315.)}
\label{tab:overlap}
\par\vspace{\baselineskip}
\begin{tabular*}{\linewidth}{@{\extracolsep{\fill}}lcccc@{}}
\toprule
Evaluation set & Tablets & Text in Stage-A & Verbatim lines & Strict subset \\
\midrule
Development (Ur~III; selection) & \Tnum{3.0}{50} & \rev{\Tfraction{3}{1}{50}\rlap{$^\S$}} & \Tnum{2.1}{34.6}\% & \Tnum{2.1}{49.8}\% \\
In-period (Ur~III; reused)           & \Tnum{3.0}{100} & \Tfraction{3}{0}{100} & \Tnum{2.1}{48.6}\% & \Tnum{2.1}{43.2}\% \\
Cross-period                         & \Tnum{3.0}{192} & \rev{\Tfraction{3}{1}{192}\rlap{$^\dagger$}} & \Tnum{2.1}{9.0}\% & \Tnum{2.1}{84.4}\% \\
Cross-collection (HT)                & \Tnum{3.0}{315} & \Tfraction{3}{0}{282}\rlap{$^\ddagger$} & \Tnum{2.1}{1.2}\% & \Tnum{2.1}{98.0}\% \\
\bottomrule
\end{tabular*}\par\vspace{5pt}\begin{minipage}{\linewidth}\small
$^\dagger$HS~1032 \rev{(CDLI P134344, cataloged as HS~1032a/b; 15 Stage-A lines) and $^\S$HS~2445 (P101855; 4 lines)
were missed by our tablet-to-text mapping. Removing HS~1032 and HS~2080 (Appendix~\ref{app:ht}) from the
cross-period set changes the contrasts in Table~\ref{tab:main} and their strict-subset versions (Table~\ref{tab:strictser}) by at most 0.5\,pp. Other cross-period results in this paper also include both tablets.} $^\ddagger$315 manifest tablets contain 282 CDLI P-numbers; primary CTC scoring loads qualifying lines from 305 of these tablets.
\end{minipage}
\end{table}

\begin{table}[H]
\centering
\footnotesize
\caption{\textbf{Strict-subset SER ablation} (with vs.\ without Stage-A), 3-seed corpus-SER
mean over strict lines only.  Strict = exact-unseen $\cap$ no-shared-4-gram against the
Stage-A corpus, computed on each model's loaded SER records.
The \vlm evaluator loads 718/1{,}254/1{,}408 lines across sets, versus 670/1{,}243/1{,}320 for CTC,
so its strict $n$ is larger. These are the same line sets as in Table~\ref{tab:main}.  $\Delta$SER = SER$_{\text{no}}-$SER$_{\text{yes}}$; Exact = exact-line
accuracy (\%).  The strict removal cost reproduces the all-line $\Delta$ of
Table~\ref{tab:main} within $\sim$2\,pp on all nine $N$=100 contrasts (max 1.9\,pp); at
$N$=25 it departs more, by up to $\sim$5.8\,pp in-period (\svtr) and less off-domain, as the
strict subset drops the easier formulaic lines.  Source:
\texttt{compute\_strict\_ablation\_ser.py} (records only; no retrain).}
\label{tab:strictser}
\par\vspace{\baselineskip}
\begin{tabular*}{\linewidth}{@{\extracolsep{\fill}}llcccccc@{}}
\toprule
& & & & \multicolumn{2}{c}{SER (\%$\downarrow$)} & \multicolumn{2}{c}{Exact (\%$\uparrow$)} \\
\cmidrule(lr){5-6}\cmidrule(lr){7-8}
Set & Model & $N$ & $n$ & with & no & with & no \\
\midrule
\multirow{6}{*}{In-period}
& \svtr & \Tnum{3.0}{25} & \Tnum{4.0}{291} & \Tnum{2.1}{33.0} & \Tnum{3.1}{95.1} & \Tnum{2.1}{22.5} & \Tnum{2.1}{0.0} \\
& \svtr & \Tnum{3.0}{100} & \Tnum{4.0}{291} & \Tnum{2.1}{25.4} & \Tnum{3.1}{76.2} & \Tnum{2.1}{34.9} & \Tnum{2.1}{0.6} \\
& \crnn & \Tnum{3.0}{25} & \Tnum{4.0}{291} & \Tnum{2.1}{38.3} & \Tnum{3.1}{100.4} & \Tnum{2.1}{16.5} & \Tnum{2.1}{0.0} \\
& \crnn & \Tnum{3.0}{100} & \Tnum{4.0}{291} & \Tnum{2.1}{29.4} & \Tnum{3.1}{76.1} & \Tnum{2.1}{26.6} & \Tnum{2.1}{0.1} \\
& \vlm  & \Tnum{3.0}{25} & \Tnum{4.0}{386} & \Tnum{2.1}{31.9} & \Tnum{3.1}{78.7} & \Tnum{2.1}{26.9} & \Tnum{2.1}{2.4} \\
& \vlm  & \Tnum{3.0}{100} & \Tnum{4.0}{386} & \Tnum{2.1}{20.5} & \Tnum{3.1}{50.4} & \Tnum{2.1}{48.7} & \Tnum{2.1}{12.4} \\
\midrule
\multirow{6}{*}{Cross-period}
& \svtr & \Tnum{3.0}{25} & \Tnum{4.0}{1050} & \Tnum{2.1}{39.3} & \Tnum{3.1}{98.2} & \Tnum{2.1}{14.3} & \Tnum{2.1}{0.0} \\
& \svtr & \Tnum{3.0}{100} & \Tnum{4.0}{1050} & \Tnum{2.1}{40.0} & \Tnum{3.1}{91.7} & \Tnum{2.1}{14.4} & \Tnum{2.1}{0.1} \\
& \crnn & \Tnum{3.0}{25} & \Tnum{4.0}{1050} & \Tnum{2.1}{46.2} & \Tnum{3.1}{100.3} & \Tnum{2.1}{8.1} & \Tnum{2.1}{0.0} \\
& \crnn & \Tnum{3.0}{100} & \Tnum{4.0}{1050} & \Tnum{2.1}{43.9} & \Tnum{3.1}{91.5} & \Tnum{2.1}{10.4} & \Tnum{2.1}{0.0} \\
& \vlm  & \Tnum{3.0}{25} & \Tnum{4.0}{1071} & \Tnum{2.1}{40.9} & \Tnum{3.1}{94.0} & \Tnum{2.1}{14.4} & \Tnum{2.1}{0.0} \\
& \vlm  & \Tnum{3.0}{100} & \Tnum{4.0}{1071} & \Tnum{2.1}{43.0} & \Tnum{3.1}{77.3} & \Tnum{2.1}{12.0} & \Tnum{2.1}{0.7} \\
\midrule
\multirow{6}{*}{Cross-collection}
& \svtr & \Tnum{3.0}{25} & \Tnum{4.0}{1293} & \Tnum{2.1}{70.0} & \Tnum{3.1}{97.8} & \Tnum{2.1}{2.1} & \Tnum{2.1}{0.0} \\
& \svtr & \Tnum{3.0}{100} & \Tnum{4.0}{1293} & \Tnum{2.1}{71.4} & \Tnum{3.1}{98.0} & \Tnum{2.1}{1.7} & \Tnum{2.1}{0.0} \\
& \crnn & \Tnum{3.0}{25} & \Tnum{4.0}{1293} & \Tnum{2.1}{72.7} & \Tnum{3.1}{101.8} & \Tnum{2.1}{1.1} & \Tnum{2.1}{0.0} \\
& \crnn & \Tnum{3.0}{100} & \Tnum{4.0}{1293} & \Tnum{2.1}{73.8} & \Tnum{3.1}{97.5} & \Tnum{2.1}{1.1} & \Tnum{2.1}{0.0} \\
& \vlm  & \Tnum{3.0}{25} & \Tnum{4.0}{1385} & \Tnum{2.1}{65.2} & \Tnum{3.1}{97.7} & \Tnum{2.1}{3.0} & \Tnum{2.1}{0.0} \\
& \vlm  & \Tnum{3.0}{100} & \Tnum{4.0}{1385} & \Tnum{2.1}{66.3} & \Tnum{3.1}{90.1} & \Tnum{2.1}{2.6} & \Tnum{2.1}{0.0} \\
\bottomrule
\end{tabular*}
\end{table}

\begin{table}[H]
\centering
\footnotesize
\setlength{\tabcolsep}{3.5pt}
\caption{\textbf{Auxiliary LCS-Recall companion to the SER text-exposure audit}
(\S\ref{sec:overlap}; the primary strict-subset SER result is in Table~\ref{tab:strictser}).
Overlap-controlled ablation at $N$=100 (3-seed mean): \lcs and exact-line
accuracy (\%) with vs.\ without Stage-A on the \emph{strict} subset of lines that
both share no 4-gram with any Stage-A line \emph{and} have no exact verbatim
match in it (the exact-unseen condition removes short verbatim lines that a
4-gram filter alone would admit; \S\ref{sec:overlap}), next to the all-line \lcs gap.
$n$ = strict-subset lines (subsets defined on the CTC token mapping; on HT the
\vlm also scores the two subset lines the CTC evaluator fails to load, hence
$n$=1{,}295); ``all lines'' = the all-line \lcs gap in this auxiliary metric.  $N$=25
strict \lcs gaps: in-period $+$56.1/$+$59.6/$+$41.4,
cross-period $+$54.0/$+$52.4/$+$47.8, cross-collection $+$26.9/$+$27.7/$+$27.9\,pp
(\svtr/\crnn/\vlm).}
\label{tab:strict}
\par\vspace{\baselineskip}
\begin{tabular*}{\linewidth}{@{\extracolsep{\fill}}llccccccc@{}}
\toprule
& & & \multicolumn{2}{c}{with Stage-A} & \multicolumn{2}{c}{no Stage-A} & \multicolumn{2}{c}{$\Delta$LCS (pp)} \\
\cmidrule(lr){4-5}\cmidrule(lr){6-7}\cmidrule(lr){8-9}
Set & Model & $n$ & LCS & Exact & LCS & Exact & strict & all lines \\
\midrule
\multirow{3}{*}{In-period}
& \svtr & \Tnum{4.0}{290} & \Tnum{2.1}{75.3} & \Tnum{2.1}{35.1} & \Tnum{2.1}{28.5} & \Tnum{2.1}{0.6} & \Tnum{+2.1}{+46.8} & \Tnum{+2.1}{+46.7} \\
& \crnn & \Tnum{4.0}{290} & \Tnum{2.1}{71.6} & \Tnum{2.1}{26.7} & \Tnum{2.1}{28.5} & \Tnum{2.1}{0.1} & \Tnum{+2.1}{+43.1} & \Tnum{+2.1}{+44.0} \\
& \vlm  & \Tnum{4.0}{290} & \Tnum{2.1}{78.3} & \Tnum{2.1}{44.7} & \Tnum{2.1}{51.4} & \Tnum{2.1}{11.3} & \Tnum{+2.1}{+26.9} & \Tnum{+2.1}{+26.7} \\
\midrule
\multirow{3}{*}{Cross-period}
& \svtr & \Tnum{4.0}{1049} & \Tnum{2.1}{60.4} & \Tnum{2.1}{14.5} & \Tnum{2.1}{13.5} & \Tnum{2.1}{0.1} & \Tnum{+2.1}{+46.9} & \Tnum{+2.1}{+47.7} \\
& \crnn & \Tnum{4.0}{1049} & \Tnum{2.1}{56.8} & \Tnum{2.1}{10.5} & \Tnum{2.1}{11.9} & \Tnum{2.1}{0.0} & \Tnum{+2.1}{+44.9} & \Tnum{+2.1}{+45.8} \\
& \vlm  & \Tnum{4.0}{1049} & \Tnum{2.1}{57.2} & \Tnum{2.1}{11.5} & \Tnum{2.1}{24.8} & \Tnum{2.1}{0.6} & \Tnum{+2.1}{+32.4} & \Tnum{+2.1}{+33.0} \\
\midrule
\multirow{3}{*}{Cross-collection}
& \svtr & \Tnum{4.0}{1293} & \Tnum{2.1}{30.0} & \Tnum{2.1}{1.7} & \Tnum{2.1}{5.9} & \Tnum{2.1}{0.0} & \Tnum{+2.1}{+24.1} & \Tnum{+2.1}{+24.1} \\
& \crnn & \Tnum{4.0}{1293} & \Tnum{2.1}{27.7} & \Tnum{2.1}{1.1} & \Tnum{2.1}{5.3} & \Tnum{2.1}{0.0} & \Tnum{+2.1}{+22.4} & \Tnum{+2.1}{+22.5} \\
& \vlm  & \Tnum{4.0}{1295} & \Tnum{2.1}{34.5} & \Tnum{2.1}{2.4} & \Tnum{2.1}{12.5} & \Tnum{2.1}{0.0} & \Tnum{+2.1}{+22.0} & \Tnum{+2.1}{+22.0} \\
\bottomrule
\end{tabular*}
\end{table}

\paragraph{Strict-subset definitions and counts.}  Two strict computations appear in this
paper, differing only in tokenization and line universe, not in rule.

\textbf{(i)~Annotation
audit.}  The audit (\texttt{audit\_dedup.py}\,$\to$\,\texttt{AUDIT\_DEDUP.json}) normalizes
sign tokens and parses 671 in-period lines ($\geq$3 signs); 290 (43.2\%) are strict---no
verbatim match \emph{and} no shared 4-gram with the Stage-A corpus.  This computation
underlies the auxiliary LCS view (Table~\ref{tab:strict}, $n$=290).

\textbf{(ii)~SER
evaluation records.}  The SER strict ablation (Table~\ref{tab:strictser}) and the
parent-matched contrast (\S\ref{sec:strict}) use the per-line SER records with raw sign
names.  Of the 670 loaded in-period CTC lines, 480 (72\%) share no 4-gram,
345 (51\%) never recur verbatim, and 291 satisfy both conditions.
The no-4-gram set includes 189 short ($<$4-sign) verbatim recurrences.
The \vlm evaluator loads more lines (718 in-period), giving a larger strict subset
(Table~\ref{tab:strictser}).

Both computations draw strict membership from the same Stage-A corpus
(93{,}440 \emph{distinct} sign sequences after deduplication, from 387{,}315 training
instances); its 285 sign types cover 100/99.95/96.7\% of in-period/cross-period/HT evaluation
tokens.  The CTC subset sizes are 290/1{,}049/1{,}293 (annotation audit, $\geq$3-sign
line universe) vs.\ 291/1{,}050/1{,}293 (SER-record ablation, all loaded CTC lines), the
in-period/cross-period differences coming from tokenization and the line universe; the
\vlm SER-record subsets are larger on its own line universe (Table~\ref{tab:strictser}).
Each analysis retains its own filtering pipeline and reports its subset size.  The
strict-subset SER ablation (Table~\ref{tab:strictser}) keys strict membership on the unique
per-line identifier (tablet-side stem and line id); the main table (Table~\ref{tab:main}) and
the parent-matched contrast (Figure~\ref{fig:strict}) do not use that subset key---they
aggregate the per-line records directly.  Verbatim recurrence is driven by
administrative formulae: the receipt formula \texttt{ŠU BA TI} alone is 58 of the 326
recurring in-period lines and occurs 10{,}762 times in the Stage-A corpus.

\paragraph{Auxiliary LCS view (Table~\ref{tab:strict}).}  The strict-subset LCS gaps track
the SER conclusion: within 1\,pp of the all-line \lcs gaps at $N$=100 and within 5\,pp at
$N$=25.  On the strict subset the no-Stage-A in-period \lcs is 28.5/28.5/51.4\%
(\svtr/\crnn/\vlm)---low but not zero---against with-Stage-A 75.3/71.6/78.3\%.  With-Stage-A
absolute in-period \lcs is up to 6\,pp higher on the full set than on the strict subset, a
descriptive difference consistent with formula recurrence rather than a measured causal
contribution.

\paragraph{Earlier and text-disjoint checkpoint comparison.}
We compare the earlier text-exposed checkpoints against the text-disjoint retrain under the
identical in-period protocol.  In auxiliary LCS Recall, the differences are mixed and small: \svtr scores 0.3--0.7\,pp
\emph{lower} (71.9/78.3 vs.\ 72.2/79.0 at $N$=25/100), \crnn 1.9--4.0\,pp higher
(71.1/77.9 vs.\ 67.1/76.0), and CTC $N$=0 moves by
$\leq$2\,pp.  The comparison absorbs Stage-A run variation and, for \crnn, a
different DEV-selected LR, so it describes cross-generation differences
($\leq$4\,pp) and does not isolate a text-exposure effect.  \VLMTEXTEXPOSURE

\section{Cross-Collection (MaiCuBeDa-HT) Manifest and Crop-Dictionary Provenance}
\label{app:ht}

MaiCuBeDa-HT \citep{ahlborn2025maicubedaht} releases, for the MSII rendering of each
tablet side, W3C sign annotations (transliteration, line number, character index,
polygon or box, CDLI P-number link) and cropped images of the annotated lines.  We checked each released line image against the final bounding box of its cumulative
line annotation, in the sign boxes' pixel frame. Of 2{,}842 lines, 2{,}832 match
to $\le$1\,px; we drop the 10 mismatches and 36 lines without an image.
We then express each sign box relative to its line image and use the same
4-px-padded crop and stitching function as for Ur~III. Padding is clipped at
the line-image border.  Signs are ordered by character index; tokens outside the 749-label
inventory (18.5\%) are dropped from image and label under each family's token
mapping, exactly as on the Ur~III sets (the VLM mapping additionally maps numeral
signs such as \texttt{1(disz)}, hence 1{,}408 vs.\ 1{,}322 manifest lines).  Two of the
1{,}322 CTC manifest lines fail to load, leaving 1{,}320 scored lines; this is separate from
the mapping difference.  The frozen
manifest (315 tablets, 282 CDLI P-numbers, none of which occurs in the Stage-A
corpus) is hashed below.  The released per-sign
\texttt{char/} images are fixed 250$\times$250 context windows rather than sign
boxes and were not used.

\paragraph{Crop-dictionary provenance and leakage audits.}
The 199{,}243-image dictionary comprises 146{,}296 crops from the eBL annotated sign-image
dataset \citep{lewenstein2026signdataset}, 35{,}947 from an expert-annotated Old Babylonian
sign dataset \citep{smidt2026obsigns} (photorealistic renderings), and 17{,}000
augmentation variants of rare-sign crops.  The variants are produced by \texttt{build\_clean\_dict.py}
for under-represented sign types by resampling existing real crops under rotation, scaling,
brightness/contrast, blur, and additive noise---standard geometric/photometric augmentation,
not a skeleton-alignment or generative model; they add sample variety for existing classes only.
Where the eBL metadata
\citep{lewenstein2026signdataset} records a script
period, the labels are: Unknown 144{,}221; Neo-Babylonian 494; Neo-Assyrian
376; Middle Assyrian 254; Old Assyrian 235; Middle Babylonian 206; Ur~III 128;
Parthian 108; Late Babylonian 100; Hellenistic 90; Persian 36; Presargonic 27;
Fara 16; Sargonic 5.  The OB source \citep{smidt2026obsigns} is entirely Old
Babylonian; augmented crops inherit unlabeled origins.  For the tablet-overlap audit, we parsed 10{,}780 unique tablet/museum IDs from crop filenames.
We compared these against all MaiCuBeDa tablets used here: 214 training, 50 development,
100 in-period, and 192 cross-period tablets (556 unique HS numbers), plus the
315-tablet HT manifest. Both numeric HS-number and normalized full-string matching
find \emph{no overlap}. The dictionary includes 308 other Hilprecht tablets;
the separation is at tablet level. \rev{At the text level, one cross-period tablet, HS~2080, shares its text
(CDLI P503175) with its envelope HS~2357, whose 138 Old Babylonian crops are in the dictionary and occur in
2{,}809 Stage-A lines; no other dictionary object shares a CDLI text with the 556 tablets.} The three empty inventory entries (\texttt{du},
\texttt{šar\textsubscript{2}}, \texttt{600}) held only MaiCuBeDa Hilprecht crops, which were
removed when the dictionary was built; as noted in \S\ref{sec:compositional}, token
normalization never requests them, and none occurs in the Stage-A manifest or any evaluation
reference.
The sign detector's training data (450 train / 50 val tablets) shares no
tablet with the 100 in-period test tablets, which are exactly its held-out
split; the detector pool does, however, contain the Stage-B pool, DEV, and
cross-period tablets---the \rev{primary} automated evaluation (\S\ref{sec:e2e}) therefore
\rev{uses} the in-period test \rev{tablets, and the cross-period whole-side check (Appendix~\ref{app:sourcedisjoint})
shares this detector exposure.}

\begin{figure}[H]
\centering
\includegraphics[width=\linewidth]{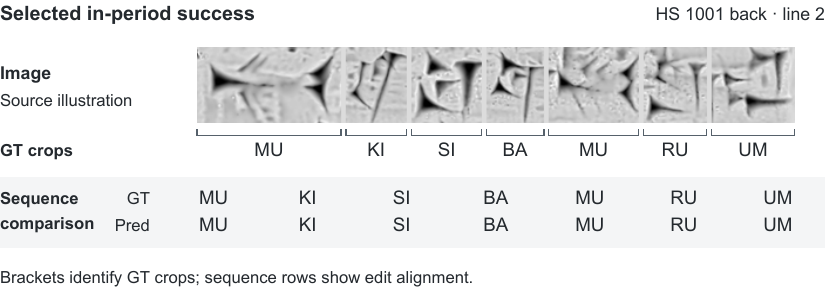}
\caption{\textbf{Selected successful in-period example} (text-disjoint \svtr, $N{=}100$, seed~42; HS~1001 back, line~2; strict subset). GT and prediction match. The GT-crop stitch illustrates the source, not the inference tensor. Brackets locate GT crops; sequence rows show edit alignment, not image positions. Rates: Table~\ref{tab:main}; strict subset: \S\ref{sec:overlap}; oracle/automatic and cross-collection examples: Figure~\ref{fig:qualitative_diagnostics}.}
\label{fig:qualitative}
\end{figure}

\begin{figure}[H]
\centering
\includegraphics[width=\linewidth]{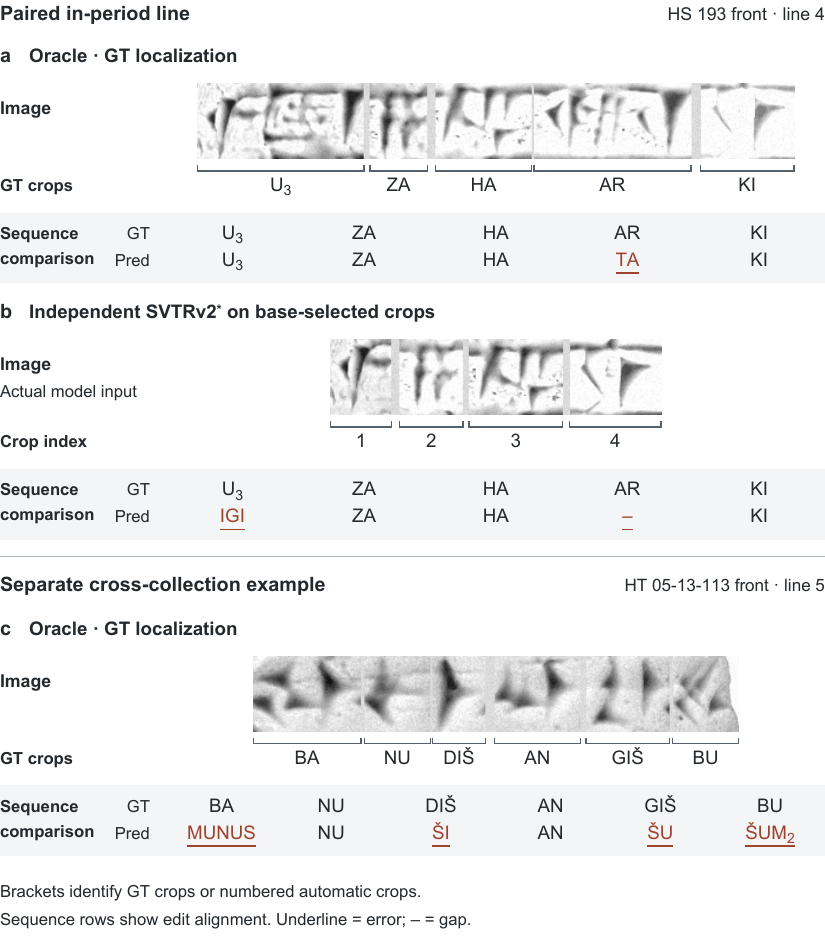}
\caption{\textbf{Localization and cross-collection examples} (text-disjoint \svtr, $N{=}100$, seed~42). \textbf{(a,b)} HS~193 front, line~4: oracle input and the exact automatic model input formed from four base-selected boxes. Panel \textbf{(b)} shows independent \svtr recognition on base-selected crops, not the full system. \textbf{(c)} An oracle-localized HT example, HT~05-13-113 front, line~5. Brackets identify GT crops in (a,c) and numbered automatic crops in (b). GT is used only for display and evaluation; sequence rows show sign-token edit alignment, not image positions (ties: diagonal, deletion, insertion). These selected cases do not represent overall rates (Tables~\ref{tab:main} and~\ref{tab:auto-shared-controls}).}
\label{fig:qualitative_diagnostics}
\label{fig:qualitative_ht}%
\end{figure}

\section{Training and Compute Details}
\label{app:training}

\paragraph{Evaluation determinism.}  Within each matched comparison and input protocol,
checkpoints are evaluated on fixed images (fixed augmentation seed), and CTC decoding is
verified against the training-time evaluator on CPU for each evaluation batch.

\paragraph{Reproduction entry points.} In the code package,
\rev{\texttt{records/verify\_current.py}} is the single entry point for primary and strict-subset
SER, attribution, converged-scratch and per-sign comparisons, development controls,
and the unified automatic comparison. It checks hashes and recomputes edit distances,
LCS, aggregates, and paired tablet intervals from saved records using Python and NumPy.
\texttt{SOURCE\_PROVENANCE.json} maps the supplied records to their sources;
\texttt{SHA256SUMS.txt} binds the package files. This \rev{script verifies saved outputs; runnable
synthesis, training, and inference entry points are listed in the package README (Reproducibility Statement).}

\paragraph{Training-run disclosures.} The VLM's two-stage execution is detailed in
\S\ref{app:recognizer_recipe}. Stage-B VLM runs use gradient checkpointing and a per-process CUDA memory
cap so that two runs share one GPU; both are memory-only (the loss is bitwise
identical to the plain path; gradients agree to within GPU nondeterminism).

\paragraph{Artifact hashes.}  Analysis plan SHA-256 \texttt{1f80a787\ldots};
locked evaluation plan \texttt{27e7c053\ldots}; cross-period manifest
\texttt{e09d11d1\ldots};
deduplicated Stage-A manifest \texttt{5450bd27\ldots}; HT manifest
\texttt{f6e632bc\ldots}.  The file name, purpose, and full 64-character hash of each of these
five artifacts are listed in the code package's \texttt{ARTIFACT\_INDEX.json}.
These hashes identify the original internal artifacts.

\paragraph{Compute.}  All runs used a single NVIDIA RTX~5090 (32\,GB; CUDA 12.8).  A
primary Stage-B fine-tune takes minutes for the CTC models (9 and 17\,min at $N$=100, seed
42, for \svtr and \crnn) and ${\sim}5.6$\,h for the VLM (LoRA; run two per GPU under a 0.36
per-process memory fraction).  These are measured per-run times from the training logs, not
a whole-project total; a complete accounting of Stage-A pre-training and detector-training
compute was not preserved and is not reported.

\begin{table}[htbp]
\centering
\footnotesize
\setlength{\tabcolsep}{4pt}
\caption{\textbf{Training configuration per run type}, read from the per-run
records, training scripts, logs, and LoRA adapter configs (not generic defaults; sources and
VLM Stage-A boundaries are given in the text below).  Optimizer is AdamW throughout; weight
decay is $10^{-4}$ for the CTC runs, while the VLM run record stores none.  Gradient-norm
clip is 5.0 (CTC) / 1.0 (VLM).  CTC line images are height-normalized to 64\,px (aspect
preserved) with worker-level augmentation (inter-sign gaps, noise).  ``Budget'' is the epoch
cap, ``Stopping'' gives the DEV early-stopping patience, and ``Ckpt sel.'' is the checkpoint-selection rule; LRs
marked ``DEV-sel'' are chosen per model$\times$$N$ on DEV, while Stage-A pre-training LRs are the
script constants.  All Stage-B/ladder/scratch runs use 3 seeds $\{$42,123,456$\}$; Stage-A
pre-training uses \rev{seed 42, and the attribution models are repeated with seeds 123/456 (Appendix~\ref{app:stageaseeds})}.}
\label{tab:trainconfig}
\par\vspace{\baselineskip}
\begin{tabular}{@{}>{\raggedright\arraybackslash}p{2.6cm}l>{\raggedright\arraybackslash}p{1.95cm}clll@{}}
\toprule
Run type & LR & Scheduler & Batch & Budget & Stopping & Ckpt sel. \\
\midrule
Stage-A \svtr            & 3$\times$10$^{-4}$ & warmup 1k, cosine (0.05 floor) & 32 & 30 ep & patience 5 & best Val-SER \\
Stage-A \crnn            & 3$\times$10$^{-4}$ & cosine ($T$=30) & 32 & 30 ep  & patience 5 & best Val-SER \\
Stage-B CTC (\svtr, \crnn) & DEV-sel          & cosine ($T$=100) & 8 & $\le$100 ep & patience 15 & best DEV-\lcs \\
Stage-B ladder / parent-matched (\svtr) & DEV-sel & cosine ($T$=100) & 8 & $\le$100 ep & patience 15 & best DEV-\lcs \\
Scratch (\svtr, no St.-A) & 3$\times$10$^{-4}$ & cosine ($T$=300) & 8 & 300 ep & off (converged) & best DEV-\lcs \\
Stage-B VLM (LoRA $r$16/$\alpha$32) & DEV-sel & cosine & 8$\times$1 & 40 ep & patience 10 & best val \lcs \\
\bottomrule
\end{tabular}
\end{table}

\paragraph{Configuration sources and boundaries.}  The CTC values are read from each run's
\texttt{result.json}; the VLM Stage-B values from its \texttt{config.json} (batch 8,
accumulation 1, 40 epochs, cosine), with early stopping at patience 10 read from
\texttt{train.log} and best-val-LCS-Recall selection from \texttt{best\_meta.json} (its
\texttt{config.json} stores neither).  The DEV-selected LR grids are
$\{5\times10^{-5},10^{-4},3\times10^{-4}\}$ (CTC) and $\{5\times10^{-5},10^{-4},2\times10^{-4}\}$
(VLM), selected by development-set \lcs at seed 42; the VLM reused its original DEV-selected
values after the text-disjoint retraining, while the CTC grids were fully re-run.  The ladder/parent-matched rows are the Stage-B fine-tunes on those arms; the ladder
\emph{Stage-A} pre-training runs are fixed at 30 epochs / final checkpoint
(Appendix~\ref{app:shuf}), not the DEV-best rule of the table.  The VLM LoRA adapter is $r$16/$\alpha$32, dropout 0.05, q/k/v/o
projections (5.49\,M trainable parameters, 0.60\%), base \texttt{PaddlePaddle/PaddleOCR-VL-1.6}.

\input{sections/recognizer_recipe.tex}

\input{sections/automatic_whole_side.tex}

\section{Small-Budget Results and Per-Seed Values}
\label{app:perseed}

\begin{table}[H]
\centering\footnotesize
\setlength{\tabcolsep}{4pt}
\caption{\textbf{Stage-A benefit with 25 adaptation tablets.} SER$\downarrow$ (\%),
three-seed means; protocol and model-specific line universes as in Table~\ref{tab:main}.
$\Delta$ = SER without $-$ with Stage-A, with conditional paired-tablet 95\% CIs.
$^\dagger$Without-Stage-A \crnn seeds 123/456 selected no checkpoint and enter the
means as empty outputs (SER=100\%); seed 42 has a decoded, near-blank checkpoint.}
\label{tab:smallbudget}
\par\vspace{\baselineskip}
\begin{tabular*}{\linewidth}{@{\extracolsep{\fill}}llccc@{}}
\toprule

Set & Model & with St.-A & no St.-A & $\Delta$ (pp) [95\% CI] \\
\midrule
\multirow{3}{*}{\shortstack[l]{In-period}}
& \svtr & \Tnum{2.1}{28.4} & \Tnum{3.1}{96.3} & \Tci{+2.1}{+67.9}{65.6}{70.2} \\
& \crnn{}$^{\dagger}$ & \Tnum{2.1}{33.6} & \Tnum{3.1}{100.5} & \Tci{+2.1}{+66.9}{64.5}{69.3} \\
& \vlm  & \Tnum{2.1}{26.0} & \Tnum{3.1}{76.3} & \Tci{+2.1}{+50.2}{47.9}{52.5} \\
\midrule
\multirow{3}{*}{\shortstack[l]{Cross-period}}
& \svtr & \Tnum{2.1}{38.0} & \Tnum{3.1}{97.9} & \Tci{+2.1}{+59.9}{56.3}{62.8} \\
& \crnn{}$^{\dagger}$ & \Tnum{2.1}{44.6} & \Tnum{3.1}{100.3} & \Tci{+2.1}{+55.8}{52.1}{58.7} \\
& \vlm  & \Tnum{2.1}{39.0} & \Tnum{3.1}{93.3} & \Tci{+2.1}{+54.3}{51.4}{56.7} \\
\midrule
\multirow{3}{*}{\shortstack[l]{Cross-collection}}
& \svtr & \Tnum{2.1}{70.0} & \Tnum{3.1}{97.8} & \Tci{+2.1}{+27.8}{25.9}{29.8} \\
& \crnn{}$^{\dagger}$ & \Tnum{2.1}{72.4} & \Tnum{3.1}{101.8} & \Tci{+2.1}{+29.4}{27.5}{31.4} \\
& \vlm  & \Tnum{2.1}{64.9} & \Tnum{3.1}{97.7} & \Tci{+2.1}{+32.8}{31.2}{34.5} \\
\bottomrule
\end{tabular*}
\end{table}

At $N$=0, \vlm and \svtr are within 1\,pp SER on every set; \crnn trails by up to 7.6\,pp.
At $N$=100, the ordering varies by domain. These are descriptive comparisons:
pre-training, capacity, token mapping, and Stage-A budget differ.
Tables~\ref{tab:perseed_lr} and~\ref{tab:perseed} list every run behind
Tables~\ref{tab:main} and~\ref{tab:smallbudget}: Table~\ref{tab:perseed_lr} gives the DEV-selected learning rate of each
arm, and Table~\ref{tab:perseed} the per-seed SER.  As in \S\ref{sec:setup}, the primary test
for each contrast is whether its per-comparison 95\% tablet-bootstrap CI excludes zero (all
18 do).  Zero of the 10{,}000 bootstrap resamples reversed the sign; this empirical count is
not a calibrated $p$-value bound, and we do not rely on it in place of the CI.
\textbf{Without-Stage-A convergence.}  Most without-Stage-A CTC runs with a DEV-selected
checkpoint early-stopped on DEV well before the 100-epoch budget (best epochs 1--66; the two
failed \crnn $N$=25 runs below record best epoch 0, a sentinel for ``no checkpoint
selected''); the \crnn $N$=100 runs stopped latest (best epochs 87/82/97; two of the three
reached the 100-epoch budget).  As a
converged control we retrain the \svtr $N{=}100$ no-Stage-A arm
from scratch for 300 epochs (early stop off, cosine over the full budget; training loss
saturates near zero and DEV LCS peaks by epoch 244--267 under this schedule); it
narrows the removal cost by only 1.5--4.2\,pp---to $+$46.7 [44.5,\,48.9],
$+$50.7 [47.7,\,53.1], and $+$25.2 [23.4,\,27.0]\,pp in-, cross-period, and
cross-collection (3-seed mean, tablet bootstrap 95\% CI).  The \svtr advantage thus persists under the tested 300-epoch budget.
For without-Stage-A \crnn at $N$=25, seeds 123/456 kept DEV LCS at 0.
Training stopped at epoch 15 with loss ${\approx}$4.2 and no selected checkpoint.
Their records are explicit failures scored as empty outputs (all deletions, SER $=$ 100\%).
Seed 42 selected epoch 46 of 61 (loss 3.25); its near-blank output scores 101.5\% in-period.  This is an optimization failure under the fixed budget and grid,
and the affected Table~\ref{tab:smallbudget} rows are marked ($\dagger$).  The
Stage-A effect is thus measured at a fixed downstream annotation and Stage-B budget; the
conclusions rest on the $N$=100 contrasts (24--52\,pp SER), not on the degenerate cells. An additional
fixed-100-epoch control revisits these runs in Appendix~\ref{app:additional_checks}.

\begin{table}[htbp]
\centering
\footnotesize
\setlength{\tabcolsep}{5pt}
\caption{\textbf{DEV-selected learning rates} (with / without Stage-A) per model and $N$, chosen by
seed-42 development \lcs and applied to all three seeds and all evaluation sets (the rate does
not vary by set).  These are the LRs behind the per-seed SER in Table~\ref{tab:perseed}.}
\label{tab:perseed_lr}
\par\vspace{\baselineskip}
\begin{tabular*}{\linewidth}{@{\extracolsep{\fill}}lccc@{}}
\toprule
Model & $N$ & LR with Stage-A & LR without Stage-A \\
\midrule
\svtr & \Tnum{3.0}{25} & 10$^{-4}$          & 10$^{-4}$ \\
\svtr & \Tnum{3.0}{100} & 3$\times$10$^{-4}$ & 3$\times$10$^{-4}$ \\
\crnn & \Tnum{3.0}{25} & 10$^{-4}$          & 3$\times$10$^{-4}$ \\
\crnn & \Tnum{3.0}{100} & 3$\times$10$^{-4}$ & 3$\times$10$^{-4}$ \\
\vlm  & \Tnum{3.0}{25} & 5$\times$10$^{-5}$ & 2$\times$10$^{-4}$ \\
\vlm  & \Tnum{3.0}{100} & 2$\times$10$^{-4}$ & 2$\times$10$^{-4}$ \\
\bottomrule
\end{tabular*}
\end{table}

\begin{table}[H]
\centering
\footnotesize
\setlength{\tabcolsep}{5pt}
\caption{\textbf{Per-seed SER} (\%$\downarrow$) for the 18 primary contrasts (seeds
42/123/456); in-period = the reused set of \S\ref{sec:splits}.  DEV-selected LRs are in
Table~\ref{tab:perseed_lr}.  $\Delta$ = the SER cost of Tables~\ref{tab:main} and~\ref{tab:smallbudget} (rounded
once from full precision, so it may differ from the difference of the two rounded seed means
by $\leq$0.1\,pp); the 3-seed means reproduce Tables~\ref{tab:main} and~\ref{tab:smallbudget}.
$^{\dagger}$Without-Stage-A \crnn $N$=25: seeds 123/456 kept DEV LCS at 0 and stopped
at epoch 15 without a selected checkpoint. These failures are scored as empty outputs
(all deletions, 100.0\%). Seed 42 decodes near-blank outputs scoring 101.5/100.9/105.4\%.  Per-seed auxiliary LCS is in the per-line
records of the code package.}
\label{tab:perseed}
\par\vspace{\baselineskip}
\begin{tabular*}{\linewidth}{@{\extracolsep{\fill}}llcccc@{}}
\toprule
Set & Model & $N$ & \shortstack[l]{with Stage-A\\(s42/s123/s456)} & \shortstack[l]{without\\(s42/s123/s456)} & $\Delta$ \\
\midrule
In-period & \svtr & \Tnum{3.0}{25} & \Ttriple{2.1}{28.5}{28.5}{28.1} & \Ttriple{3.1}{97.2}{94.9}{96.7} & \Tnum{+2.1}{+67.9} \\
In-period & \svtr & \Tnum{3.0}{100} & \Ttriple{2.1}{20.5}{22.4}{21.6} & \Ttriple{3.1}{65.3}{79.2}{72.9} & \Tnum{+2.1}{+50.9} \\
In-period & \crnn{}$^{\dagger}$ & \Tnum{3.0}{25} & \Ttriple{2.1}{32.9}{33.6}{34.4} & \Ttriple{3.1}{101.5}{100.0}{100.0} & \Tnum{+2.1}{+66.9} \\
In-period & \crnn & \Tnum{3.0}{100} & \Ttriple{2.1}{24.3}{24.6}{25.5} & \Ttriple{3.1}{72.2}{77.7}{69.3} & \Tnum{+2.1}{+48.3} \\
In-period & \vlm & \Tnum{3.0}{25} & \Ttriple{2.1}{26.0}{26.3}{25.7} & \Ttriple{3.1}{77.1}{74.2}{77.4} & \Tnum{+2.1}{+50.2} \\
In-period & \vlm & \Tnum{3.0}{100} & \Ttriple{2.1}{17.6}{17.7}{18.4} & \Ttriple{3.1}{45.8}{47.7}{44.0} & \Tnum{+2.1}{+27.9} \\
\midrule
Cross-period & \svtr & \Tnum{3.0}{25} & \Ttriple{2.1}{38.2}{38.3}{37.6} & \Ttriple{3.1}{98.7}{96.0}{99.0} & \Tnum{+2.1}{+59.9} \\
Cross-period & \svtr & \Tnum{3.0}{100} & \Ttriple{2.1}{37.6}{38.4}{39.2} & \Ttriple{3.1}{87.3}{93.0}{92.1} & \Tnum{+2.1}{+52.4} \\
Cross-period & \crnn{}$^{\dagger}$ & \Tnum{3.0}{25} & \Ttriple{2.1}{44.0}{44.4}{45.3} & \Ttriple{3.1}{100.9}{100.0}{100.0} & \Tnum{+2.1}{+55.8} \\
Cross-period & \crnn & \Tnum{3.0}{100} & \Ttriple{2.1}{41.8}{42.2}{41.8} & \Ttriple{3.1}{90.7}{91.6}{89.2} & \Tnum{+2.1}{+48.6} \\
Cross-period & \vlm & \Tnum{3.0}{25} & \Ttriple{2.1}{38.8}{39.3}{38.8} & \Ttriple{3.1}{93.7}{93.1}{92.9} & \Tnum{+2.1}{+54.3} \\
Cross-period & \vlm & \Tnum{3.0}{100} & \Ttriple{2.1}{41.7}{40.9}{41.1} & \Ttriple{3.1}{75.7}{76.5}{76.5} & \Tnum{+2.1}{+35.0} \\
\midrule
Cross-collection (HT) & \svtr & \Tnum{3.0}{25} & \Ttriple{2.1}{70.1}{70.0}{69.8} & \Ttriple{3.1}{99.0}{96.6}{97.7} & \Tnum{+2.1}{+27.8} \\
Cross-collection (HT) & \svtr & \Tnum{3.0}{100} & \Ttriple{2.1}{71.1}{71.3}{71.1} & \Ttriple{3.1}{97.3}{98.3}{97.9} & \Tnum{+2.1}{+26.7} \\
Cross-collection (HT) & \crnn{}$^{\dagger}$ & \Tnum{3.0}{25} & \Ttriple{2.1}{72.1}{72.1}{73.0} & \Ttriple{3.1}{105.4}{100.0}{100.0} & \Tnum{+2.1}{+29.4} \\
Cross-collection (HT) & \crnn & \Tnum{3.0}{100} & \Ttriple{2.1}{74.6}{72.9}{73.3} & \Ttriple{3.1}{97.6}{98.2}{96.5} & \Tnum{+2.1}{+23.9} \\
Cross-collection (HT) & \vlm & \Tnum{3.0}{25} & \Ttriple{2.1}{65.4}{64.7}{64.7} & \Ttriple{3.1}{97.3}{97.4}{98.6} & \Tnum{+2.1}{+32.8} \\
Cross-collection (HT) & \vlm & \Tnum{3.0}{100} & \Ttriple{2.1}{66.2}{66.8}{65.4} & \Ttriple{3.1}{90.5}{90.4}{89.3} & \Tnum{+2.1}{+23.9} \\
\bottomrule
\end{tabular*}
\end{table}

\section{Per-Period Breakdown of the Cross-Period Set}
\label{app:perperiod}

\noindent The cross-period set spans six attested periods of very different size and
difficulty; Table~\ref{tab:perperiod} reports per-period tablet and token counts and $N$=100
performance.

\begin{table}[H]
\centering
\small
\caption{\textbf{Cross-period performance is heterogeneous:} the largest period
(ED~IIIb) sits 13--16\,pp below the micro average.  Per-period \lcs (\%) at
$N$=100 with Stage-A (3-seed mean, aggregated within period); token counts use
the CTC serialization (the VLM is scored on its own records, which add 11
lines).  Small cells warrant caution: ED~IIIa has 3 tablets and 48 tokens;
\rev{HS~1032 is Ur~III but had no period label because our catalog mapping missed it (Table~\ref{tab:overlap}).}  Approximate
dates: ED~IIIb $\sim$2500--2340\,BC; ED~IIIa and Old Akkadian also precede Ur~III,
whereas Old Assyrian and Old Babylonian follow it. The range extends up to
$\sim$500 years from Ur~III.}
\label{tab:perperiod}
\par\vspace{\baselineskip}
\begin{tabular*}{\linewidth}{@{\extracolsep{\fill}}lccccc@{}}
\toprule
Period & Tablets & Tokens & \svtr & \crnn & \vlm \\
\midrule
Old Assyrian          & \Tnum{3.0}{44} & \Tnum{4.0}{2054} & \Tnum{2.1}{65.7} & \Tnum{2.1}{62.3} & \Tnum{2.1}{63.4} \\
Old Babylonian        & \Tnum{3.0}{13} & \Tnum{4.0}{1614} & \Tnum{2.1}{69.4} & \Tnum{2.1}{64.7} & \Tnum{2.1}{60.2} \\
ED IIIb               & \Tnum{3.0}{114} & \Tnum{4.0}{1584} & \Tnum{2.1}{46.5} & \Tnum{2.1}{44.1} & \Tnum{2.1}{46.6} \\
Early Old Babylonian  & \Tnum{3.0}{7} & \Tnum{4.0}{662} & \Tnum{2.1}{72.1} & \Tnum{2.1}{70.1} & \Tnum{2.1}{73.5} \\
Old Akkadian          & \Tnum{3.0}{10} & \Tnum{4.0}{157} & \Tnum{2.1}{56.9} & \Tnum{2.1}{51.4} & \Tnum{2.1}{56.5} \\
ED IIIa               & \Tnum{3.0}{3} & \Tnum{4.0}{48} & \Tnum{2.1}{33.3} & \Tnum{2.1}{30.6} & \Tnum{2.1}{40.3} \\
\rev{Ur~III (HS~1032)}      & \Tnum{3.0}{1} & \Tnum{4.0}{38} & \Tnum{2.1}{81.6} & \Tnum{2.1}{81.6} & \Tnum{2.1}{89.7} \\
\midrule
Macro avg.\ (6 periods) & {---} & {---} & \Tnum{2.1}{57.3} & \Tnum{2.1}{53.9} & \Tnum{2.1}{56.7} \\
Micro avg.\ (all)       & \Tnum{3.0}{192} & \Tnum{4.0}{6157} & \Tnum{2.1}{62.0} & \Tnum{2.1}{58.7} & \Tnum{2.1}{59.3} \\
\bottomrule
\end{tabular*}
\end{table}

\section{Attribution Study: Protocol, Sequence-Order Control, and Seed Stability}
\label{app:shuf}

\noindent This appendix gives the full attribution protocol, the all-lines $N{=}100$ SER
reductions (Table~\ref{tab:ladder_panelb}), the parent-matched construction, and the separate
global-shuffle sensitivity analysis.

\begin{table}[H]\centering\small
\caption{\textbf{Three-arm attribution study} (\svtr, per-arm SER$\downarrow$ \%; \svtr is our
SVTRv2-inspired CTC variant, \S\ref{sec:models}).
Per-arm absolute SER on each set for the isolated-sign, shuffled-order, and attested-order
Stage-A arms.  Lifting isolated-sign supervision to multi-sign line format gives the largest tested
training-format gain on every set.  $N{=}100$ values are 3-seed Stage-B means; the isolated-sign arm is an
initialization/task-transfer control.  Corresponding all-lines reductions with CIs are in
Table~\ref{tab:ladder_panelb}; the strict, parent-matched contrasts in
Figure~\ref{fig:strict}.}
\label{tab:ladder}
\par\vspace{\baselineskip}
\begin{tabular*}{\linewidth}{@{\extracolsep{\fill}}lcccccc@{}}
\toprule
 & \multicolumn{2}{c}{in-period} & \multicolumn{2}{c}{cross-period} & \multicolumn{2}{c}{cross-collection}\\
\cmidrule(lr){2-3}\cmidrule(lr){4-5}\cmidrule(lr){6-7}
Stage-A arm & $N{=}0$ & $N{=}100$ & $N{=}0$ & $N{=}100$ & $N{=}0$ & $N{=}100$\\
\midrule
Isolated-sign (control) & \Tnum{2.1}{95.4} & \Tnum{2.1}{61.8} & \Tnum{2.1}{97.5} & \Tnum{2.1}{82.3} & \Tnum{2.1}{99.0} & \Tnum{2.1}{94.6} \\
Shuffled-order line      & \Tnum{2.1}{36.4} & \Tnum{2.1}{25.6} & \Tnum{2.1}{39.2} & \Tnum{2.1}{38.6} & \Tnum{2.1}{68.3} & \Tnum{2.1}{71.2} \\
Attested-order line      & \Tnum{2.1}{32.8} & \Tnum{2.1}{21.2} & \Tnum{2.1}{39.2} & \Tnum{2.1}{34.4} & \Tnum{2.1}{68.2} & \Tnum{2.1}{69.5} \\
\bottomrule
\end{tabular*}
\end{table}

\begin{table}[H]\centering\small
\caption{\textbf{All-lines $N{=}100$ SER reductions} (pp) [95\% CI] for the three-arm study
(\svtr\ attribution family; 3-seed Stage-B mean; paired tablet bootstrap, conditional on the
trained Stage-A checkpoints).  These are the all-lines contrasts corresponding to
Table~\ref{tab:ladder}; the strict, parent-matched contrasts are in
Figure~\ref{fig:strict}.}
\label{tab:ladder_panelb}
\par\vspace{\baselineskip}
\begin{tabular}{l S[table-format=2.1]@{\ [}S[table-format=2.1]@{,\,}S[table-format=2.1]@{]\hspace{2\tabcolsep}} S[table-format=2.1]@{\ [}S[table-format=2.1]@{,\,}S[table-format=2.1]@{]\hspace{2\tabcolsep}} S[table-format=2.1]@{\ [}S[table-format=2.1]@{,\,}S[table-format=2.1]@{]}}
\toprule
Reduction & \multicolumn{3}{c}{in-period} & \multicolumn{3}{c}{cross-period} & \multicolumn{3}{c}{\makebox[0pt]{cross-collection}}\\
\midrule
isolated$\to$shuffled & 36.2 & 34.7 & 37.7 & 43.7 & 40.5 & 46.7 & 23.5 & 21.6 & 25.4\\
shuffled$\to$attested & 4.4 & 3.3 & 5.5 & 4.2 & 3.0 & 5.2 & 1.7 & 0.8 & 2.7\\
isolated$\to$attested & 40.6 & 38.9 & 42.3 & 47.9 & 44.3 & 51.1 & 25.2 & 23.3 & 27.1\\
\bottomrule
\end{tabular}
\end{table}

\textbf{Primary three-arm protocol (Table~\ref{tab:ladder}).}  All three arms share
one on-the-fly renderer and corpus (text-disjoint: 387{,}315 lines, 1{,}778{,}825
(sign, crop) tokens, 285 types, 146{,}784 distinct crops; height 64, gaps 2--8\,px,
vertical jitter $\pm3$, Gaussian noise $\sigma{=}6$, brightness 0.90--1.10, background
value 220, real-background padding).  \emph{Attested-order} renders each source line in
corpus order; \emph{shuffled-order} permutes all (sign, crop) tokens within each
P-number split and re-cuts them into lines of the original lengths; \emph{isolated-sign}
renders each token as its own one-sign image.  A \emph{parent} is the group of tokens the
loss treats as one unit for grouping and weighting: in the line arms one parent is one
rendered line, and in the isolated-sign arm the tokens of a parent are rendered as separate
one-sign images but still weighted together as that parent.  Every arm uses a batch of 32
parents per optimizer step with a parent-grouped CTC loss (mean over the tokens of a parent,
then mean over the 32 parents), so per-parent weighting is identical across arms; in the
isolated-sign arm each parent group yields ${\approx}4.59$ one-sign images, so that arm sees
${\approx}147$ images per step versus 32 line-images for the line arms.  Token exposure and optimizer updates per epoch are matched; total compute
is \emph{not} claimed equal.  Each arm trains for a fixed 30 epochs (no early stopping;
the epoch-30 checkpoint is used), Stage-A seed 42, AdamW (LR $3\times10^{-4}$, weight
decay $10^{-4}$, 1000-step warmup then cosine), 10\% P-number validation split.  Stage-B
then follows the locked protocol (LR chosen from $\{5{\times}10^{-5},10^{-4},
3{\times}10^{-4}\}$ by seed-42 DEV LCS, applied to seeds 123/456; $N\in\{25,100\}$; \svtr
only).  The all-lines reduction CIs (Table~\ref{tab:ladder_panelb}) and the strict CIs
(Figure~\ref{fig:strict}) are tablet bootstraps conditional on these single
Stage-A checkpoints and therefore exclude Stage-A training \rev{variance, which Appendix~\ref{app:stageaseeds}
estimates with two further seeds.}  The two-seed stability check below concerns the separate
global-shuffle control family (a distinct, more aggressive \rev{perturbation).}

\textbf{Design boundary and parent-matched control.}  The isolated-to-shuffled contrast
also changes token grouping: the shuffled arm re-groups tokens into new lines, whereas
the isolated-sign arm keeps single tokens.  So on its own that step only evaluates the
\emph{bundled} multi-sign line-format change (canvas, target length, CTC alignment,
grouping, and multi-sign exposure jointly), not stitching in isolation.  We therefore
add a \emph{parent-matched} control that renders the \emph{same} shuffled groups as
isolated one-sign images, holding grouping fixed (isolated vs.\ multi-sign line-format inputs; canvas, length, and CTC training still co-vary):
singleton-matched $\to$ shuffled still lowers SER by $+$45.1/$+$46.2/$+$23.1\,pp on the
strict subset in-period, cross-period, and cross-collection, respectively (3-seed mean,
tablet bootstrap; \S\ref{sec:strict}).  The gain therefore persists when token grouping is held fixed.

\begin{table}[H]\centering\small
\caption{\textbf{Strict-subset attribution: full values} plotted in Figure~\ref{fig:strict}
(\svtr\ attribution family; $N{=}100$; 3-seed Stage-B mean; SER-record strict subset,
exact-unseen $\cap$ no-shared-4-gram; paired tablet bootstrap 95\% CI, conditional on the
Stage-A checkpoints).  SER reductions (pp); positive $=$ improvement.  The first contrast is
parent-matched (grouping held fixed); the second is not; the point differences telescope through the shared shuffled arm, but their CIs cannot
simply be added. They do not provide a causal decomposition of the main Stage-A effect.}
\label{tab:strictci}
\par\vspace{\baselineskip}
\begin{tabular*}{\linewidth}{@{\extracolsep{\fill}}lccc@{}}
\toprule
Contrast & in-period & cross-period & cross-collection\\
\midrule
singleton-matched $\to$ shuffled & \Tci{+2.1}{45.1}{42.5}{47.7} & \Tci{2.1}{46.2}{42.8}{49.4} & \Tci{2.1}{23.1}{21.2}{25.1} \\
shuffled $\to$ attested          & \Tci{+2.1}{1.2}{-0.7}{3.0} & \Tci{2.1}{3.5}{2.1}{4.8} & \Tci{2.1}{1.8}{0.9}{2.8} \\
\bottomrule
\end{tabular*}
\end{table}

\begin{table}[h]
\centering
\footnotesize
\setlength{\tabcolsep}{3.5pt}
\caption{\textbf{Global-shuffle order control (\svtr, \lcs \%).}  A stronger
perturbation than the within-split shuffle of the primary attribution study
(Table~\ref{tab:ladder}): Stage-A on the real-order text-disjoint corpus vs.\ the
same corpus with (sign, crop) pairs \emph{globally} shuffled.  $N$=25/100 are
3-seed Stage-B means (locked protocol; DEV-selected LRs $10^{-4}$/$3\times10^{-4}$
for both), $N$=0 the Stage-A models alone.  $\Delta$ = real $-$ shuffled, 95\%
tablet-bootstrap CI (seed stability below; $N$=0 $\Delta$: single checkpoints, no
CI).  Real order is worth ${\sim}4$\,pp in-period; under this stronger perturbation
the off-domain effect is smaller and seed-dependent (cf.\ the milder within-split shuffle, \rev{whose in-period benefit is positive for all three Stage-A seeds;
Table~\ref{tab:ladder}, Appendix~\ref{app:stageaseeds}}).}
\label{tab:shuf}
\par\vspace{\baselineskip}
\begin{tabular*}{\linewidth}{@{\extracolsep{\fill}}lccccccccc@{}}
\toprule
& \multicolumn{3}{c}{$N$=0} & \multicolumn{3}{c}{$N$=25} & \multicolumn{3}{c}{$N$=100} \\
\cmidrule(lr){2-4}\cmidrule(lr){5-7}\cmidrule(lr){8-10}
Set & real & shuf. & $\Delta$ & real & shuf. & $\Delta$ [CI] & real & shuf. & $\Delta$ [CI] \\
\midrule
In-period (Ur~III) & \Tnum{2.1}{66.8} & \Tnum{2.1}{62.5} & \Tnum{+1.1}{+4.3} & \Tnum{2.1}{72.2} & \Tnum{2.1}{68.6} & \Tci{+1.1}{+3.7}{2.0}{5.4} & \Tnum{2.1}{79.0} & \Tnum{2.1}{74.9} & \Tci{+1.1}{+4.1}{2.9}{5.3} \\
Cross-period       & \Tnum{2.1}{59.7} & \Tnum{2.1}{61.1} & \Tnum{+1.1}{-1.4} & \Tnum{2.1}{62.4} & \Tnum{2.1}{64.6} & \Tci{+1.1}{-2.1}{-3.4}{-0.9} & \Tnum{2.1}{62.0} & \Tnum{2.1}{62.5} & \Tci{+1.1}{-0.5}{-1.7}{0.7} \\
Cross-collection   & \Tnum{2.1}{30.7} & \Tnum{2.1}{33.2} & \Tnum{+1.1}{-2.4} & \Tnum{2.1}{31.5} & \Tnum{2.1}{34.4} & \Tci{+1.1}{-2.9}{-3.8}{-1.9} & \Tnum{2.1}{30.2} & \Tnum{2.1}{29.8} & \Tci{+1.1}{+0.4}{-0.4}{1.3} \\
\bottomrule
\end{tabular*}
\end{table}

\textbf{Separate global-shuffle control.}  This separate sensitivity analysis is distinct
from the primary within-split shuffle of Table~\ref{tab:ladder}.  From the text-disjoint
corpus (387{,}315 lines, 1{,}778{,}825 sign tokens, 285 types, 146{,}784 distinct crop
images) we list every (sign, crop) pair in corpus order, apply one global permutation
(seed 42), and re-cut the permuted list into lines of the original lengths, attaching to
each line the source line's P-number.  Verified equalities with the real-order corpus: line
count, token count, per-sign counts, crop multiset, (sign, crop)-pair multiset, and
line-length distribution are identical.  Each shuffled line keeps a source-line P-number
\emph{label} only for bookkeeping; because tokens are permuted \emph{globally across splits},
this does not preserve the original per-line (sign, crop) membership of the validation split
(unlike the primary within-split shuffle of Table~\ref{tab:ladder}, which is unaffected).
The renderer (height 64, gaps 2--8\,px, jitter, noise,
brightness) and the Stage-A/Stage-B training configurations are unchanged.  2{,}243
shuffled lines (0.58\%) coincide with some real line; 2/3/4-gram overlap with the real
corpus is 51.6/7.9/0.4\%, and the shuffled corpus has 1.8/5.1/4.5$\times$ more
distinct 2/3/4-grams.  Manifest SHA-256 \texttt{d835d53f\ldots}.
\textbf{Seed stability of the global-shuffle control.}  \SHUFSEEDS

\section{VLM Diagnostics: Full Tables}
\label{app:vlm}

\noindent This appendix collects two \vlm diagnostics: an image-sensitivity probe
(Table~\ref{tab:imgsens}) and the LoRA-rank check.  (The cross-generation text-exposure
comparison is in Appendix~\ref{app:strict}.)

\begin{table}[htbp]
\centering
\small
\setlength{\tabcolsep}{6pt}
\caption{\textbf{Image-sensitivity results} on the 670-line in-period set (earlier \vlm
         Stage-A checkpoint; the three Stage-B checkpoints are summarized in the text).
         Permuted: images randomly reassigned (derangement, seed=99).
         Blank: solid gray (pixel 128).
         Noise: Gaussian $\mathcal{N}(128,50^2)$.
         ChgRate: \% of lines where prediction differs from Correct.
         The harness re-stitches the input images itself (fresh gap/noise draw), so
         Correct-row \lcs can differ slightly from the same legacy checkpoints' scores.}
\label{tab:imgsens}
\par\vspace{\baselineskip}
\begin{tabular*}{\linewidth}{@{\extracolsep{\fill}}llccc@{}}
\toprule
Checkpoint & Condition & LCS & Exact\% & ChgRate\% \\
\midrule
\multirow{4}{*}{\vlm\ Stage-A ($N$=0)}
& Correct    & \Tnum{2.2}{73.26} & \Tnum{2.2}{41.49} & \Tnum{2.1}{0.0} \\
& Permuted   & \Tnum{2.2}{7.29} & \Tnum{2.2}{1.49} & \Tnum{2.1}{98.5} \\
& Blank      & \Tnum{2.2}{6.39} & \Tnum{2.2}{0.75} & \Tnum{2.1}{99.1} \\
& Noise      & \Tnum{2.2}{6.92} & \Tnum{2.2}{0.60} & \Tnum{2.1}{99.3} \\
\bottomrule
\end{tabular*}
\end{table}

The three Stage-B checkpoints evaluated ($r$=16 at $N$=100 and $N$=200;
$r$=64 at $N$=100) behave the same way: permuted, blank, and noise inputs
change 98.7--100\% of predictions and collapse \lcs to 0.71--7.93\%
(Correct rows: 75.13--81.90\%).

\paragraph{LoRA rank.}  At $N$=100 under matched tablet subsets (seeds 123 and
456), we see no systematic rank effect across the two controlled seeds:
$r$=16/32/64 reach auxiliary LCS Recall of $72.27\pm0.82$, $72.25\pm1.11$, and $71.95\pm0.74\%$
(mean$\pm$std over the two seeds).  With only two seeds this is a small stability
check, not an equivalence test.  The seed-42 runs (76.07, 81.94, 82.72\%) reused the
original production tablet subset, so their larger spread is confounded with subset
selection.

\FloatBarrier
{\revon\input{sections/natural_matched.tex}}

\FloatBarrier
\section{Strong Per-Sign Classifier Baseline}
\label{app:persign}

Because the oracle protocol supplies GT sign boxes, we compare \svtr against a strong per-sign
classifier that keeps the protocol but replaces sequence decoding with per-crop classification
(main-text summary in \S\ref{sec:crossperiod}).

\paragraph{Configuration and sources.}
\textbf{Architecture / initialization:} ImageNet-pre-trained DeiT3-medium
(\texttt{deit3\_medium\_patch16\_224.fb\_in22k\_ft\_in1k}, ImageNet-22k$\to$1k)
with a linear head over 749 labels (sorted dictionary signs $+$ \texttt{X} $+$ \texttt{lacuna}).
It uses no MAE, multitask head, or earlier cuneiform weights.
\textbf{Stage-A:} dictionary classification on 199{,}243 crops
(\texttt{dictionary\_clean.json}, tablet-disjoint from all test sets).
The run config records 193{,}266 training and 5{,}977 validation crops
($\texttt{val\_frac}{=}0.03$). Training uses AdamW, LR $10^{-4}$, weight decay 0.05,
cosine scheduling, and at most 30 epochs with patience 6.
Best validation per-sign accuracy is 0.860 at epoch 25.
\textbf{Stage-B:} adaptation uses the main table's 100-tablet \texttt{train\_n100} split
and seeds 42/123/456. As for \svtr, checkpoints maximize LCS Recall on the 50-tablet DEV set.
The saved DEV curves confirm patience 8. No per-run training manifest or config was saved;
the split and AdamW LR $10^{-5}$ come from the retained implementation, with the LR
unconfirmed by a run record.

\textbf{Inference / scoring:} classify each GT crop and concatenate predictions in line order.
SER, exact-line accuracy, and strict membership use the same GT, OOV rule, and
\texttt{(stem,line\_id)} keys as \svtr. The evaluated line sets match exactly:
670/1{,}243/1{,}320 full lines and 291/1{,}050/1{,}293 strict lines.
Saved predictions record checkpoint paths but no checkpoint hashes.
A later hash inventory is part of the code package; it does not establish a
contemporaneous checkpoint-to-output binding.
CIs are paired tablet bootstraps ($B{=}10{,}000$, percentile 95\%),
conditional on fixed checkpoints, with $\Delta$ = SER$_{\text{per-sign}}-$SER$_{\text{\svtr}}$.
Architecture, initialization, and objective differ, so this comparison does not
isolate sequence context.

\begin{table}[H]
\centering\small
\setlength{\tabcolsep}{5pt}
\caption{\textbf{Strong per-sign classifier vs.\ \svtr under the oracle protocol} ($N{=}100$,
3-seed mean).  SER$\downarrow$ and exact-line$\uparrow$ (\%); $\Delta$ =
SER$_{\text{per-sign}}-$SER$_{\text{\svtr}}$ (negative $=$ classifier lower), paired tablet-bootstrap
95\% CI conditional on the fixed checkpoints.  Per-sign top-1 accuracy: in-period 79.0,
cross-period 72.2, cross-collection 37.6.}
\label{tab:persign}
\par\vspace{\baselineskip}
\begin{tabular*}{\linewidth}{@{\extracolsep{\fill}}llccccc@{}}
\toprule
Set & Subset & SER$_{\text{p-s}}$ & SER$_{\text{\svtr}}$ & $\Delta$ (pp) [95\% CI] & Ex$_{\text{p-s}}$ & Ex$_{\text{\svtr}}$ \\
\midrule
\multirow{2}{*}{In-period} & full   & \Tnum{2.1}{21.0} & \Tnum{2.1}{21.5} & \Tci{+2.1}{-0.5}{-1.8}{0.9} & \Tnum{2.1}{44.9} & \Tnum{2.1}{47.1} \\
& strict & \Tnum{2.1}{22.2} & \Tnum{2.1}{25.4} & \Tci{+2.1}{-3.2}{-5.5}{-0.8} & \Tnum{2.1}{39.6} & \Tnum{2.1}{34.9} \\
\midrule
\multirow{2}{*}{Cross-period} & full   & \Tnum{2.1}{27.8} & \Tnum{2.1}{38.4} & \Tci{+2.1}{-10.6}{-12.2}{-8.9} & \Tnum{2.1}{25.9} & \Tnum{2.1}{16.4} \\
& strict & \Tnum{2.1}{28.3} & \Tnum{2.1}{40.0} & \Tci{+2.1}{-11.7}{-13.2}{-10.0} & \Tnum{2.1}{25.3} & \Tnum{2.1}{14.4} \\
\midrule
\multirow{2}{*}{Cross-collection} & full   & \Tnum{2.1}{62.4} & \Tnum{2.1}{71.2} & \Tci{+2.1}{-8.8}{-9.9}{-7.7} & \Tnum{2.1}{3.6} & \Tnum{2.1}{1.7} \\
& strict & \Tnum{2.1}{62.5} & \Tnum{2.1}{71.4} & \Tci{+2.1}{-8.8}{-10.0}{-7.7} & \Tnum{2.1}{3.5} & \Tnum{2.1}{1.7} \\
\bottomrule
\end{tabular*}
\end{table}

On full lines the in-period difference is small with a CI that includes zero; on the in-period
strict subset and on both off-domain sets the intervals favor the per-sign classifier (CIs
excluding zero). \rev{The two are complementary: rescoring the classifier's candidates with the Stage-A \svtr
beats the classifier alone on all three sets (Appendix~\ref{app:persign_fusion}), and \svtr is far cheaper to run
on CPU (Appendix~\ref{app:speed}).} Automatic recognition is evaluated separately in
Appendix~\ref{app:e2edetail}.

\input{sections/development_checks.tex}
{\revon\input{sections/added_appendix.tex}}

\end{document}

%% file: sections/table_typography.tex
\usepackage{siunitx}
\newcommand{\Tnum}[2]{\mbox{\tablenum[table-format=#1]{#2}}}
\newcommand{\Tci}[4]{\Tnum{#1}{#2}\,[\Tnum{#1}{#3},\,\Tnum{#1}{#4}]}
\newcommand{\Tinterval}[3]{[\Tnum{#1}{#2},\,\Tnum{#1}{#3}]}
\newcommand{\Ttriple}[4]{\Tnum{#1}{#2}\,/\,\Tnum{#1}{#3}\,/\,\Tnum{#1}{#4}}
\newcommand{\Tfraction}[3]{\mbox{#2/\makebox[1.5em][r]{#3}}}

%% file: tables/automatic_comparison.tex
\begin{table}[htbp]
\centering\small
\caption{\textbf{Original geometric-route baselines and full system.}
Mean whole-side SER over three recognition-seed runs on the 100 reused in-period tablets;
all outputs are scored. 95\% CIs condition on fixed models. Budgets and decoding differ
(Table~\ref{tab:auto-resources}; Appendix~\ref{app:e2edetail}).}
\label{tab:auto-whole-side}
\par\vspace{\baselineskip}
\begin{tabular*}{\linewidth}{@{\extracolsep{\fill}}lccc@{}}
\toprule
System & Runs & SER (\%$\downarrow$) & 95\% CI \\
\midrule
Geometric + \crnn & \Tnum{1.0}{3} & \Tnum{2.2}{72.22} & \Tinterval{2.2}{68.64}{76.12} \\
Geometric + \svtr & \Tnum{1.0}{3} & \Tnum{2.2}{70.78} & \Tinterval{2.2}{67.26}{74.63} \\
Geometric + \vlm & \Tnum{1.0}{3} & \Tnum{2.2}{68.92} & \Tinterval{2.2}{65.42}{72.81} \\
\addlinespace[4pt]
\textbf{Full system} & \Tnum{1.0}{3} & \textbf{\Tnum{2.2}{38.05}} & \Tinterval{2.2}{35.82}{40.43} \\
\bottomrule
\end{tabular*}
\end{table}

%% file: sections/recognizer_recipe.tex
\subsection{Recognizer definitions and serialization}
\label{app:recognizer_recipe}

\paragraph{CTC architectures.}
The retained implementations use grayscale $64\times W$ tensors in $[0,1]$.
Both convolutional stems have four $3\times3$, stride-1, padding-1 convolutions
with channels $1\!\to\!64\!\to\!128\!\to\!256\!\to\!256$, each followed by
batch normalization, ReLU, and max pooling. Pool sizes/strides are
$(2,2),(2,2),(2,1),(2,1)$, yielding height 4 and width $\lfloor W/4\rfloor$.
CRNN averages over height, then applies two bidirectional LSTM layers with
256 hidden units per direction and inter-layer dropout 0.2; a
$512\!\to\!750$ linear head produces CTC logits (3,975,534 trainable parameters).
For \svtr, a learned query attends over the four height positions independently
at each column (256 dimensions, four heads), followed by layer normalization.
Sinusoidal positional encoding precedes six residual mixing/FFN blocks in the
order \texttt{LLGLLG}: L is a kernel-3 depthwise 1D convolution with GELU;
G is four-head self-attention. Both use pre-layer normalization, and each FFN
is $256\!\to\!512\!\to\!256$ with GELU and dropout 0.1; global-attention dropout
is also 0.1. Final layer normalization and a $256\!\to\!750$ head give
3,532,398 parameters. There is no semantic-guidance module.

\paragraph{Cropping, batching, and decoding.}
Real-line inputs use 4-pixel-padded sign boxes, Lanczos height resizing,
and corner-median background normalization to intensity 220 (invert first
if the estimated background is below 80), followed by 2--8-pixel gaps and
Gaussian noise with $\sigma=6$. CTC minibatches are right-padded with zero.
Stage-A CTC losses use each image's $\lfloor W/4\rfloor$ valid length;
Stage-A \svtr{} also masks padded positions in global attention.
The retained Stage-B implementations differ: \svtr{} uses valid CTC lengths
but passes no attention padding mask, whereas CRNN uses the full padded
output length for each batch member. These conventions are shared by the
with/without-Stage-A arms within each model. Primary CTC loss uses mean
reduction and \texttt{zero\_infinity=True}; greedy decoding collapses consecutive
repeated classes, then removes blank. The separately controlled attribution
renderer and parent-weighted objective are specified in Appendix~\ref{app:shuf}.

\paragraph{Inventory and canonical mapping.}
Class order is the sorted 747 dictionary keys followed by \texttt{X} and
\texttt{lacuna}; blank has index 749. Of those keys, 744 have crop images.
The code package's \rev{\texttt{records/core/labels/inventory.json}} and
\texttt{token\_mapping.py} preserve the class order and alias map.
Mapping handles bracketed reconstructions as \texttt{lacuna}, illegible
\texttt{X}, editorial markers, ASCII transliteration spellings, numeric
subscripts, case, and aliases before inventory lookup. Both mappings reject
pure numeral strings; CTC additionally rejects numeral-prefix expressions
such as \texttt{1(disz)}, whereas the VLM can retain them when mapped to an
inventory entry. The VLM also rejects normalized strings longer than
24 characters. Thus the model families have different scored line universes.

\paragraph{VLM prompting and output.}
The model and its processor/tokenizer are loaded from
\texttt{PaddlePaddle/PaddleOCR-VL-1.6}. A user message contains the RGB image
and the following fixed prompt, also used outside Ur~III:
\begin{quote}\small
Transliterate this Ur III Sumerian cuneiform line.
Output only space-separated sign names (e.g. BA AN KI).
Do not explain. Do not translate. Do not output natural language.
\end{quote}
The processor's chat template supplies the generation prefix.
Training appends tokenized, space-separated sign names and EOS; prompt
positions have loss label $-100$, so only response tokens contribute.
Oracle evaluation uses \texttt{do\_sample=False} and at most 64 new
model-token IDs. Its historical code decodes
\texttt{output[prompt\_length:-1]}, unconditionally omitting the last generated
ID, then applies the saved parser: remove recognized response prefixes,
split at whitespace or \texttt{-/,;}, normalize signs, and retain only
inventory entries. Oracle SER is computed after this parsing.
Automatic evaluation scores full outputs (Appendix~\ref{app:e2edetail}).

\paragraph{Stage-A records and reproduction limits.}
The CTC logs record 347,322 training and 39,993 validation pairs, split by
P-number from the 387,315-pair corpus; selected epochs are 30 (\svtr) and
23 (CRNN). The VLM's initial executed configuration has 126,000 training
records, batch 8, seed 42, LR $2\times10^{-4}$, and a three-epoch schedule.
The selected lineage continues the saved step-42,000 adapter for 5,250
batch-8 updates on 42,000 records, with fresh AdamW/cosine states at
$2.3\times10^{-5}$; this is not uninterrupted replay of that initial schedule.
No per-run sampled-ID manifest or optimizer/RNG continuation state was
retained; the run records do not pin a processor/tokenizer revision. VLM weight decay 0.01 is specified
by the retained training code but not its run configuration.
The \rev{code package's \texttt{records/} folder} supplies saved outputs, label mappings, split identifiers,
and a minimal Stage-A text manifest for independent overlap checks; \rev{the package excludes
images and weights.} The missing run states prevent exact replay of historical training.

%% file: sections/automatic_whole_side.tex
\section{Automatic Whole-Side Recognition}
\label{app:e2edetail}

\subsection{Inputs, supervision, and frozen components}
\label{app:auto_refinement}
The proposed modular pipeline maps a complete tablet-side image and its surface type
(front, back, or edge) to an ordered sequence of canonical sign labels.
Reference boxes, reference lines, and reference text are not inputs.
Table~\ref{tab:auto-resources} separates the supervision of its components.
The detector uses 450 training and 50 validation tablets; these contain, respectively,
43 and 7 of the recognizer DEV tablets, but none of the 100 evaluation tablets.
The \rev{primary} automatic experiment therefore \rev{uses} the in-period evaluation \rev{set, because} the detector pool also
contains cross-period \rev{tablets; Appendix~\ref{app:sourcedisjoint} adds a cross-period check and a source-disjoint set.}

\input{tables/automatic_resources.tex}

YOLO26n \citep{jocher2026yolo26} produces proposals at confidence $0.05$, NMS IoU $0.90$, image size $960$,
and at most $1{,}000$ detections. Nonfinite or nonpositive-area boxes are rejected
and numeric duplicates collapsed. A fixed detector calibrator supplies proposal
probabilities. Geometric line hypotheses and a conservative, training-derived
column rule determine output order. For detected two-column layouts, the supplied
surface type gives left-to-right column order on fronts and right-to-left order on
backs; for other surfaces, including edges, geometric row order is retained.
These rules do not read reference annotations.

The DeiT3-medium classifier \citep{touvron2022deit3} uses dictionary-classification pre-training, followed
by adaptation on the 214-tablet pool (seeds 123/42/456). Of these tablets, 210 contribute
7,582 usable in-inventory characters, including characters without a line label.
AdamW uses learning rate $10^{-5}$, weight decay $10^{-4}$, batch size 32, and a
30-epoch cosine schedule and patience 8; strict-improvement DEV LCS selection retains
epochs 26, 25, 20, respectively. The classifier uses
RGB sign crops with zero extra padding at inference and the same 749-label
inventory. This component does not use synthetic-line pre-training.
Predicted signs retain their supporting detector crops.

\subsection{Original geometric-route baselines}
\label{app:original_auto_baselines}
The original baselines use YOLO proposals, geometric line grouping, and the
text-disjoint, with-Stage-A $N{=}100$ \crnn, \svtr, or \vlm checkpoints
(seeds 42/123/456). The detector weights are identical to those of the full system,
but the original settings are confidence $0.20$ followed by a $0.25$ filter,
image size $640$, NMS IoU $0.70$, and at most 300 detections.
Grouping sorts boxes by center-$y$, splitting at successive gaps above
$\max(0.5\,\text{median box height},8)$ pixels, and orders boxes by center-$x$
within a group. There is no DeiT3-based crop selection or language-model reranking.

The completed baseline evaluation retains all 198 sides and all 1,106 predicted
clusters, without reference-line matching. The original minimum of three boxes
leaves 183 clusters with empty outputs; the other 923 are decoded.
A shared seed-42 render cache uses the original stitching recipe. Previously
matched nonempty box groups are reproduced (637 groups), but expanding traversal
to all clusters changes random draws: this is a protocol-aligned re-evaluation,
not a bitwise replay of historical matched-line scores.
CTC decoding is greedy. The original VLM has a 128-new-token limit, compared with
64 on base-selected crops. One original VLM output (seed 123) reaches its limit
and is retained; full continuations and unknown tokens are scored.
These settings and the original DEV-selected checkpoints were fixed before the
completed evaluation; the detector threshold had previously been examined on this
historically reused evaluation set. The comparison is consequently not a new blind
test or a controlled estimate of changing only the input front end.

\subsection{Independent recognition on base-selected crops}
\label{app:improved_inputs}
The independent \crnn, \svtr, and \vlm comparisons reuse the
text-disjoint, with-Stage-A $N{=}100$ checkpoints for seeds 42/123/456.
They receive the same crops and order selected by the \emph{base} per-sign decoder,
not by the candidate verifier, and do not receive DeiT3's sign labels.
Stitches use height 64, crop padding 4, gaps of 2--8 pixels, the original noise
recipe, and a deterministic seed per proposal path. Input hashes agree across
all nine runs; the VLM additionally converts grayscale to RGB.

The independent recognizers process all 95 DEV sides (50 tablets) and 198 evaluation
sides (100 tablets).
Their 696 and 1,186 proposed rows, including 18 and 78 empty rows, are retained.
No output rows are dropped by matching to reference lines.
CTC recognizers retain greedy decoding. VLM outputs retain unknown tokens and
trailing content after canonical sign-name normalization; none of the three
evaluation runs reaches the 64-new-token generation limit.
Oracle-line results use the vocabulary-filtered protocol in Appendix~\ref{app:recognizer_recipe}.

\subsection{Candidate construction and sequence verification}
\label{app:svtr_verifier}
The base pool combines the top five visual classes, beam size 64, and up to 16
hypotheses per generator, using 17 fixed weight vectors with either no language
model or a six-gram modified Kneser--Ney model,
queried with KenLM \citep{heafield2011kenlm,heafield2013scalable}.
The LM uses the text-disjoint CDLI Ur~III corpus \citep{cdli2023}:
deduplicating exact sequences within each P-number leaves 332,876 lines and
1,528,710 tokens from 71,929 P-numbers. Sources exclude \rev{the} P-numbers linked
to DEV, the in-period evaluation set, \rev{and the cross-period test, except two texts missed by our mapping
(HS~1032, HS~2445; Table~\ref{tab:overlap});} Stage-B training texts remain eligible. Exact strings can still recur across distinct tablets.

An empty candidate is always available. Candidate generation uses
$T_{\rm gen}=1.55$; reranking recomputes visual log probabilities from raw logits
at $T_{\rm rank}=3$, without changing the pool. The base score is
\begin{equation}
\begin{split}
s_0(h)={}&\sum_{t\in h}\log p_{T_{\rm rank}}(t)
+\lambda_d D(h)+\lambda_\ell\log P_{\rm LM}(h)\\
&+\lambda_o O(h)+b|h|+a\,\mathbf{1}[|h|>0],
\end{split}
\end{equation}
where $D$ sums calibrated detector log odds, $O$ penalizes overlapping
successive proposals, and $\lambda_d$, $\lambda_\ell$, $\lambda_o$, $b$, and $a$ are the
DEV-selected base-decoder weights.

The adopted extension uses an existing compositionally pre-trained $N{=}100$
\svtr (seeds 42/123/456). Its adaptation tablets are a subset of the classifier's
214-tablet pool. Each candidate $h$ is scored on a stitch $x_h$ of its own selected
detector crops, rather than the shared base-selected stitch used by the
independent recognizers:
\begin{equation}
s(h)=s_0(h)+\lambda f(h), \qquad
f(h)=\frac{\log P_{\rm CTC}(h\mid x_h)}{|h|}.
\end{equation}
The feature is zero for an empty candidate. We use the full CTC sequence
probability \citep{graves2006ctc}, not the greedy-path probability.
Scoring neither creates boxes nor aligns outputs to reference text or CTC
emission times. Figure~\ref{fig:auto_verifier} shows this inference path.
\input{sections/automatic_flow.tex}

Only $\lambda\in\{0,0.1,0.25,0.5,1,2,4\}$ is selected for this extension.
In each of the existing five DEV tablet folds, base-decoder weights remain
fold-specific and $\lambda$ is selected on the other 40 tablets.
Under the \emph{historical annotated-ignore DEV objective}, conditional
out-of-fold errors fall from 534 to 529 over 1,769 reference signs.
The all-DEV choice is $\lambda=0.25$. With the existing 300-epoch,
without-Stage-A $N{=}100$ \svtr control, the same procedure selects $\lambda=0$
in every fold and on all \rev{DEV; the matched 100-epoch no-Stage-A model receives $\lambda=0$ in two of three runs
(Appendix~\ref{app:sourcedisjoint}).} This development check is conditional on the detector's exposure to DEV and
the DEV-selected recognition checkpoints.
The weight remains fixed when evaluation changes to full-output scoring.

\subsection{Reference construction, scoring, and uncertainty}
\label{app:auto_scoring}
All systems use one fixed reference on the same 100 tablets and 198 sides:
3,280 in-inventory signs, with 903 other annotated regions excluded because the fixed
mapper does not assign their readings to the 749-label inventory (387 are numeral notations;
the rest are readings such as \emph{babbar} and \emph{suen}).
Every parseable character accepted by the fixed mapper contributes, including
short rows and characters without a \texttt{Line} field.
Within each logical \texttt{(Column, Line)} group, complete, positive, unique
\texttt{Charindex} values determine order: a logical line can wrap across physical
rows, so sorting all characters by $x$ is insufficient \citep{homburg2025maicubeda}.
Ambiguous indices and missing-line assignments use the frozen annotation-geometry
fallback while preserving indexed sequences. Numeric column labels determine
reference column order; logical groups within a column follow geometric vertical
order. These reference-construction rules are not inference inputs.

Predicted rows are concatenated in their predicted order, retaining every output
token. There is no reference-row matching, spatial ignore masking, or deletion of
outputs near excluded annotations. SER is
$\sum_{\text{sides}}(S+D+I)/\sum_{\text{sides}}|Y|$, where $|Y|$ is the length of a side's reference.
Outputs corresponding to the 903 excluded regions can therefore incur errors,
even when they depict real signs. This is a fixed closed-inventory task, not a
full-inventory scholarly transcription score.

Intervals use 10,000 percentile bootstrap resamples of whole tablets, seed
20260923, with one shared sampling matrix for all systems and seeds.
Each row reports the arithmetic mean of three corpus-level recognition-seed SERs;
per-seed values are in Table~\ref{tab:auto-seeds}.
For the complete pipeline and its ablation, the fixed classifier/verifier seed pairs are
$(123,42)$, $(42,123)$, and $(456,456)$; each classifier regenerates its own candidates.
The two added classifiers use the original training recipe and DEV selection rule.
Both Stage-A initializations, detector, LM, and decoding parameters remain fixed;
there is no decoder retuning. These are recognition-seed repeats, not independent
retraining of every system component. Table~\ref{tab:auto-full-repeats} reports
all three pairs and their sample standard deviations.
The intervals condition on these models, excluding retraining and selection
uncertainty. Paired differences are computed at full precision before rounding.

\subsection{Overall gains, stronger controls, and module ablation}
The main comparison (Table~\ref{tab:auto-whole-side}) separates the original geometric
baselines from the final pipeline. Full-system SER reductions are 34.17\,pp relative to \crnn\ (paired 95\% CI $[31.38,37.08]$),
32.73\,pp relative to \svtr\ (paired 95\% CI $[30.08,35.61]$),
and 30.87\,pp relative to \vlm\ (paired 95\% CI $[28.10,33.88]$).
The main text uses the strongest of these original baselines, \vlm, to summarize
the overall gain; all three are reported.

\input{tables/automatic_shared_controls.tex}
As stronger controls, independent recognizers decode crops and order already
selected by the original seed-123 base per-sign decoder. The full system reduces SER by 8.34\,pp
relative to \svtr, the lowest observed SER among these three controls
(paired 95\% CI $[7.05,9.68]$; Table~\ref{tab:auto-shared-controls}).
This comparison uses an improved input selection procedure and is distinct from
the original geometric baseline comparison.
\input{tables/automatic_seeds.tex}
\FloatBarrier

\input{tables/automatic_full_repeats.tex}
\input{tables/automatic_verifier_errors.tex}
The matched ablation (Table~\ref{tab:verifier_extension}) reduces mean edit errors
from 1308.0 to 1248.0, with fewer substitutions and insertions but more
deletions. Mean output length changes from 3210.3 to 3048.3 tokens;
mean LCS recovery is 2356.7 versus 2348.3 of 3,280 reference signs, and
mean exact-side counts are 3.67 versus 3.00 of 198.
The original $(123,42)$ pair scores 37.59\% SER and supplies the qualitative
automatic example.
The comparison supports use of the
existing sequence model for candidate selection, not a general architecture
\rev{advantage; the matched with/without-Stage-A comparison is in Appendix~\ref{app:sourcedisjoint}.}

\subsection{Reproduction and interpretation}
All 18 recognizer runs (nine original-route and nine base-selected-crop controls)
retain their original DEV-selected checkpoints and complete outputs.
The code package will include predictions, reference records, shared-input
hashes, and tablet-bootstrap arrays for saved-record verification.
A raw-image DEV replay taking only an image and surface metadata reproduced
the selected candidate scores and tokens. Record replay does not retrain the
models or regenerate their neural predictions.

The \rev{in-period} evaluation set has been inspected \rev{historically; Appendix~\ref{app:sourcedisjoint} adds a
source-disjoint set.}
Current edit categories do not separately identify the contributions of detection,
crop selection, and line grouping.
The present comparison measures complete frozen systems under a shared
evaluation target, with different component supervision and decoding.

%% file: tables/automatic_resources.tex
\begin{table}[htbp]
\centering\small
\caption{\textbf{Supervision and selection of automatic components.}
Tablet counts belong to individual components and are not additive: the $N{=}100$
recognizer set is contained in the $N{=}214$ classifier pool. \rev{In-period and source-disjoint evaluation} tablets are
separate from adaptation and detector training/validation tablets. DEV was seen by
the detector and used for component/decoder selection. Inference takes an image
and surface type, without reference boxes, lines, or text.}
\label{tab:auto-resources}
\par\vspace{\baselineskip}
\begin{tabular}{@{}>{\raggedright\arraybackslash}p{0.17\linewidth}>{\raggedright\arraybackslash}p{0.36\linewidth}>{\raggedright\arraybackslash}p{0.41\linewidth}@{}}
\toprule
Component & Direct supervision & Selection and inference role \\
\midrule
YOLO26n & 450 training and 50 validation tablets with sign boxes & 43 DEV tablets in training; 7 in validation. All routes share weights; geometric and base-selected routes use different detection settings. \\
DeiT3 & Dictionary classification pre-training; adaptation on 214 tablets (7,582 usable characters from 210 tablets) & Seeds 123/42/456; the same DEV checkpoint-selection rule. Seed 123 supplies base-selected controls; all three run the final pipeline. Absent from geometric baselines. \\
Six-gram LM & 332,876 deduplicated CDLI text lines; 71,929 P-numbers & Excludes DEV/evaluation-associated \rev{P-numbers except two missed texts (Table~\ref{tab:overlap}).} Scores base-decoder/final candidates; absent from geometric baselines. \\
Sequence recognizers & Compositional Stage-A; $N{=}100$ adaptation, seeds 42/123/456 & Same DEV-selected checkpoints decode geometric or base-selected inputs without DeiT3 labels. \\
Candidate verifier & Reuses $N{=}100$ \svtr, seeds 42/123/456; no extra adaptation tablets & Final pipeline only: full CTC probability on each candidate's own stitch; $\lambda=0.25$ selected using historical DEV ignore scoring. \\
\bottomrule
\end{tabular}
\end{table}

%% file: sections/automatic_flow.tex
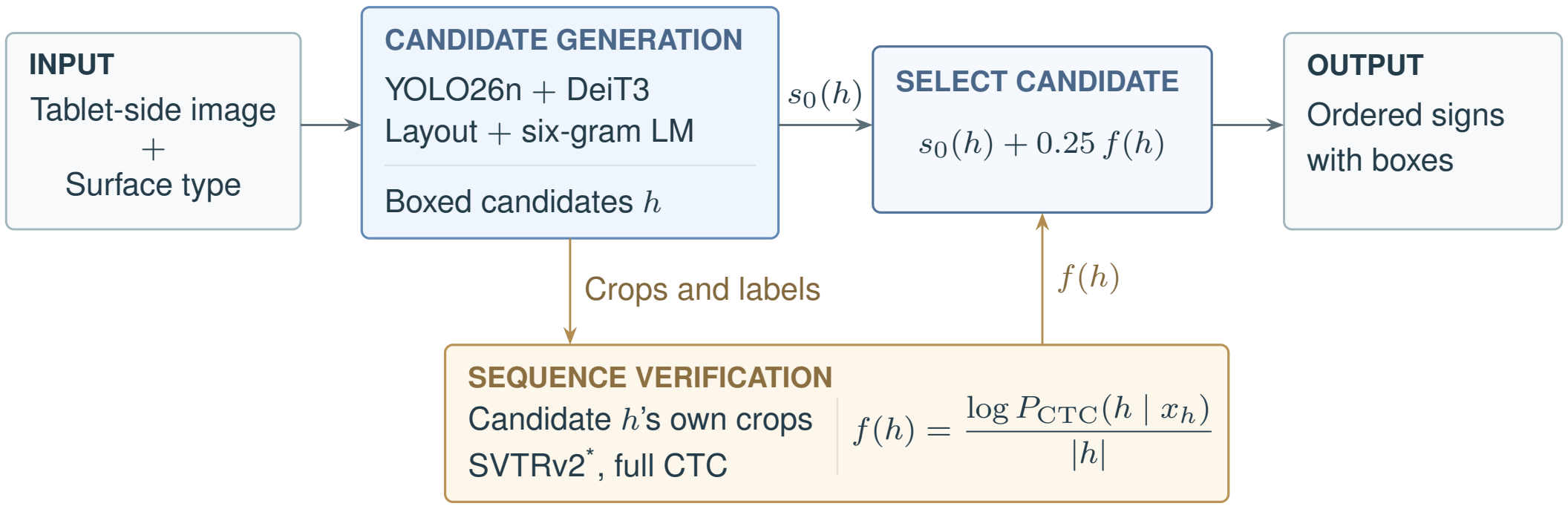
\begin{figure}[!t]
\centering
\definecolor{autotext}{HTML}{243C49}
\definecolor{autogray}{HTML}{F7F9FA}
\definecolor{autograyedge}{HTML}{A3B5BE}
\definecolor{autobluefill}{HTML}{EFF5FC}
\definecolor{autoblueedge}{HTML}{628AB7}
\definecolor{autowarmfill}{HTML}{FCF6EB}
\definecolor{autowarmedge}{HTML}{BA9457}
\begin{tikzpicture}[x=.99cm,y=1cm,
  every node/.style={text=autotext,font=\sffamily\fontsize{8}{10.5}\selectfont},
  heading/.style={anchor=west,inner sep=0pt,font=\sffamily\bfseries\fontsize{7.2}{9}\selectfont},
  body/.style={anchor=west,inner sep=0pt},
  flow/.style={-{Stealth[length=1.6mm,width=1.2mm]},draw=autotext!75,line width=.55pt},
  verifyflow/.style={flow,draw=autowarmedge!95!black}]
\path[use as bounding box] (0,0) rectangle (14.0,4.65);
\draw[rounded corners=2pt,fill=autogray,draw=autograyedge,line width=.55pt]
  (0,2.55) rectangle (2.65,4.30);
\node[heading] at (.20,4.02) {INPUT};
\node[inner sep=0pt] at (1.325,3.59) {Tablet-side image};
\node[inner sep=0pt] at (1.325,3.26) {$+$};
\node[inner sep=0pt] at (1.325,2.93) {Surface type};
\draw[rounded corners=2pt,fill=autobluefill,draw=autoblueedge,line width=.6pt]
  (3.20,2.47) rectangle (6.95,4.53);
\node[heading,text=autoblueedge!75!black] at (3.40,4.24) {CANDIDATE GENERATION};
\node[body] at (3.40,3.79) {YOLO26n $+$ DeiT3};
\node[body] at (3.40,3.40) {Layout $+$ six-gram LM};
\draw[draw=autoblueedge!25,line width=.4pt] (3.40,3.12) -- (6.75,3.12);
\node[body] at (3.40,2.80) {Boxed candidates $h$};
\draw[rounded corners=2pt,fill=autobluefill!65,draw=autoblueedge!85!black,line width=.6pt]
  (7.80,2.70) rectangle (10.85,4.18);
\node[heading,text=autoblueedge!75!black] at (8.00,3.87) {SELECT CANDIDATE};
\node at (9.325,3.30) {$s_0(h)+0.25\,f(h)$};
\draw[rounded corners=2pt,fill=autogray,draw=autograyedge,line width=.55pt]
  (11.50,2.55) rectangle (14.0,4.30);
\node[heading] at (11.70,4.02) {OUTPUT};
\node[body] at (11.70,3.55) {Ordered signs};
\node[body] at (11.70,3.17) {with boxes};
\draw[flow] (2.65,3.48) -- (3.20,3.48);
\draw[flow] (6.95,3.48) -- node[above,font=\fontsize{8}{10}\selectfont] {$s_0(h)$} (7.80,3.48);
\draw[flow] (10.85,3.48) -- (11.50,3.48);
\draw[rounded corners=2pt,fill=autowarmfill,draw=autowarmedge,line width=.6pt]
  (3.95,.12) rectangle (11.00,1.52);
\node[heading,text=autowarmedge!75!black] at (4.15,1.23) {SEQUENCE VERIFICATION};
\node[body] at (4.15,.84) {Candidate $h$'s own crops};
\node[body] at (4.15,.45) {\svtr, full CTC};
\draw[draw=autowarmedge!30,line width=.4pt] (7.48,.36) -- (7.48,1.05);
\node at (9.25,.74) {$f(h)=\dfrac{\log P_{\rm CTC}(h\mid x_h)}{|h|}$};
\draw[verifyflow] (5.075,2.47) -- node[right,text=autowarmedge!70!black,font=\sffamily\fontsize{8}{10}\selectfont] {Crops and labels} (5.075,1.52);
\draw[verifyflow] (9.325,1.52) -- node[right,text=autowarmedge!70!black,font=\fontsize{8}{10}\selectfont] {$f(h)$} (9.325,2.70);
\end{tikzpicture}
\caption{\textbf{Modular end-to-end recognition with sequence verification.}
YOLO uses 450 training/50 validation tablets, DeiT3 a 214-tablet adaptation pool,
and the verifier an included 100-tablet subset. The verifier scores each candidate's
labels on its own crops with full CTC probability; it creates no boxes or
reference alignment. The weight $0.25$ was selected using historical DEV ignore
scoring and frozen; final evaluation retains all outputs. Empty candidates have $f(h)=0$.
Inference takes only a side image and surface type. Independent recognizers
instead share the selected crops without classifier labels
(Table~\ref{tab:auto-shared-controls}).}
\label{fig:auto_verifier}
\end{figure}

%% file: tables/automatic_shared_controls.tex
\begin{table}[htbp]
\centering\small
\caption{\textbf{Strong controls using base-selected crops and order.}
All systems use the same full-output scoring target as Table~\ref{tab:auto-whole-side}.
The three recognizers receive identical base-selected crop images and order,
without classifier labels, and report three-seed means. These inputs come from
the original classifier seed 123; the full system averages three recognition-seed pairs.
These are complete-system comparisons; 95\% CIs condition on fixed models and budgets differ
(Table~\ref{tab:auto-resources}).}
\label{tab:auto-shared-controls}
\par\vspace{\baselineskip}
\begin{tabular*}{\linewidth}{@{\extracolsep{\fill}}lccc@{}}
\toprule
System & Runs & SER (\%$\downarrow$) & 95\% CI \\
\midrule
\crnn & \Tnum{1.0}{3} & \Tnum{2.2}{55.81} & \Tinterval{2.2}{52.68}{59.39} \\
\svtr & \Tnum{1.0}{3} & \Tnum{2.2}{46.39} & \Tinterval{2.2}{43.92}{49.02} \\
\vlm & \Tnum{1.0}{3} & \Tnum{2.2}{47.01} & \Tinterval{2.2}{44.20}{50.18} \\
\addlinespace[4pt]
\textbf{Full system} & \Tnum{1.0}{3} & \textbf{\Tnum{2.2}{38.05}} & \Tinterval{2.2}{35.82}{40.43} \\
\bottomrule
\end{tabular*}
\end{table}

%% file: tables/automatic_seeds.tex
\begin{table}[!htbp]
\centering\small
\caption{\textbf{Per-seed automatic recognition.}
Each run covers 198 sides and 3,280 reference signs using a DEV-selected,
text-disjoint, with-Stage-A $N{=}100$ checkpoint. Images and order are shared
within each route, not across routes. The geometric route retains 1,106 clusters
(183 too small to decode); the base-selected route retains 1,186 rows (78 empty).
No output is selected by reference-line matching. Seeds vary the recognizer,
not all components. VLM generation caps are 128 and 64 tokens, respectively.}
\label{tab:auto-seeds}
\par\vspace{\baselineskip}
\begin{tabular*}{\linewidth}{@{\extracolsep{\fill}}lccccc@{}}
\toprule
Recognizer & Seed & SER (\%$\downarrow$) & $S$ & $D$ & $I$ \\
\midrule
\multicolumn{6}{l}{\emph{Original geometric route}} \\
\crnn & \Tnum{3.0}{42} & \Tnum{2.2}{71.71} & \Tnum{4.0}{1175} & \Tnum{3.0}{175} & \Tnum{4.0}{1002} \\
\crnn & \Tnum{3.0}{123} & \Tnum{2.2}{71.89} & \Tnum{4.0}{1175} & \Tnum{3.0}{176} & \Tnum{4.0}{1007} \\
\crnn & \Tnum{3.0}{456} & \Tnum{2.2}{73.05} & \Tnum{4.0}{1184} & \Tnum{3.0}{180} & \Tnum{4.0}{1032} \\
\svtr & \Tnum{3.0}{42} & \Tnum{2.2}{69.51} & \Tnum{4.0}{1048} & \Tnum{3.0}{174} & \Tnum{4.0}{1058} \\
\svtr & \Tnum{3.0}{123} & \Tnum{2.2}{71.86} & \Tnum{4.0}{1104} & \Tnum{3.0}{169} & \Tnum{4.0}{1084} \\
\svtr & \Tnum{3.0}{456} & \Tnum{2.2}{70.98} & \Tnum{4.0}{1109} & \Tnum{3.0}{168} & \Tnum{4.0}{1051} \\
\vlm & \Tnum{3.0}{42} & \Tnum{2.2}{68.02} & \Tnum{4.0}{1050} & \Tnum{3.0}{173} & \Tnum{4.0}{1008} \\
\vlm & \Tnum{3.0}{123} & \Tnum{2.2}{69.42} & \Tnum{4.0}{1013} & \Tnum{3.0}{164} & \Tnum{4.0}{1100} \\
\vlm & \Tnum{3.0}{456} & \Tnum{2.2}{69.33} & \Tnum{4.0}{1046} & \Tnum{3.0}{159} & \Tnum{4.0}{1069} \\
\addlinespace[4pt]
\multicolumn{6}{l}{\emph{Base-selected crops and order}} \\
\crnn & \Tnum{3.0}{42} & \Tnum{2.2}{55.24} & \Tnum{4.0}{1044} & \Tnum{3.0}{260} & \Tnum{4.0}{508} \\
\crnn & \Tnum{3.0}{123} & \Tnum{2.2}{55.03} & \Tnum{4.0}{1020} & \Tnum{3.0}{275} & \Tnum{4.0}{510} \\
\crnn & \Tnum{3.0}{456} & \Tnum{2.2}{57.16} & \Tnum{4.0}{1074} & \Tnum{3.0}{242} & \Tnum{4.0}{559} \\
\svtr & \Tnum{3.0}{42} & \Tnum{2.2}{45.67} & \Tnum{4.0}{842} & \Tnum{3.0}{363} & \Tnum{4.0}{293} \\
\svtr & \Tnum{3.0}{123} & \Tnum{2.2}{46.83} & \Tnum{4.0}{879} & \Tnum{3.0}{360} & \Tnum{4.0}{297} \\
\svtr & \Tnum{3.0}{456} & \Tnum{2.2}{46.68} & \Tnum{4.0}{890} & \Tnum{3.0}{351} & \Tnum{4.0}{290} \\
\vlm & \Tnum{3.0}{42} & \Tnum{2.2}{46.92} & \Tnum{4.0}{857} & \Tnum{3.0}{318} & \Tnum{4.0}{364} \\
\vlm & \Tnum{3.0}{123} & \Tnum{2.2}{46.86} & \Tnum{4.0}{846} & \Tnum{3.0}{305} & \Tnum{4.0}{386} \\
\vlm & \Tnum{3.0}{456} & \Tnum{2.2}{47.26} & \Tnum{4.0}{862} & \Tnum{3.0}{300} & \Tnum{4.0}{388} \\
\bottomrule
\end{tabular*}
\end{table}

%% file: tables/automatic_full_repeats.tex
\begin{table}[htbp]
\centering\small
\caption{\textbf{Full-pipeline recognition-seed repeats.}
Pairs were fixed before the two additional evaluations; the first is the original run.
Each pair covers all 198 sides. DeiT3 uses the same $N{=}214$, 7,582-character
adaptation recipe; \svtr uses $N{=}100$. Detector, LM, both Stage-A initializations,
and decoder settings are fixed. Each classifier regenerates its own candidates.
$\Delta$ is the matched SER reduction from verification, in percentage points.}
\label{tab:auto-full-repeats}
\par\vspace{\baselineskip}
\begin{tabular*}{\linewidth}{@{\extracolsep{\fill}}ccccc@{}}
\toprule
DeiT3 seed & \svtr seed & Without verifier & Full SER (\%) & $\Delta$ (pp) \\
\midrule
123 & 42 & \Tnum{2.2}{39.76} & \Tnum{2.2}{37.59} & \Tnum{1.2}{2.16} \\
42 & 123 & \Tnum{2.2}{39.73} & \Tnum{2.2}{38.11} & \Tnum{1.2}{1.62} \\
456 & 456 & \Tnum{2.2}{40.15} & \Tnum{2.2}{38.45} & \Tnum{1.2}{1.71} \\
\midrule
Mean &  & \Tnum{2.2}{39.88} & \Tnum{2.2}{38.05} & \Tnum{1.2}{1.83} \\
Sample SD &  & \Tnum{2.2}{0.24} & \Tnum{2.2}{0.43} & {---} \\
\bottomrule
\end{tabular*}
\end{table}

%% file: tables/automatic_verifier_errors.tex
\begin{table}[htbp]
\centering\small
\caption{\textbf{Sequence-verification ablation, averaged over three recognition-seed pairs.}
Full-output whole-side SER uses 3,280 fixed-inventory signs on 100 reused tablets.
The two rows use the same three candidate pools; only verification is removed.
Counts are per-run means. SER falls by 1.83 percentage points
(paired-tablet 95\% CI [1.15, 2.53]).
Edit categories do not identify the responsible module.}
\label{tab:verifier_extension}
\par\vspace{\baselineskip}
\begin{tabular*}{\linewidth}{@{\extracolsep{\fill}}lcccccc@{}}
\toprule
System & SER (\%$\downarrow$) & $S$ & $D$ & $I$ & Errors & Output tokens \\
\midrule
Without verifier & \Tnum{2.2}{39.88} & \Tnum{3.1}{633.0} & \Tnum{3.1}{372.3} & \Tnum{3.1}{302.7} & \Tnum{4.1}{1308.0} & \Tnum{4.1}{3210.3} \\
Full system & \Tnum{2.2}{38.05} & \Tnum{3.1}{528.3} & \Tnum{3.1}{475.7} & \Tnum{3.1}{244.0} & \Tnum{4.1}{1248.0} & \Tnum{4.1}{3048.3} \\
\bottomrule
\end{tabular*}
\end{table}

%% file: sections/natural_matched.tex
\section{Natural-Line Recognition}
\label{app:natural_matched}\label{app:naturalline}

\begin{figure}[H]
\centering\revon
\includegraphics[width=0.82\linewidth]{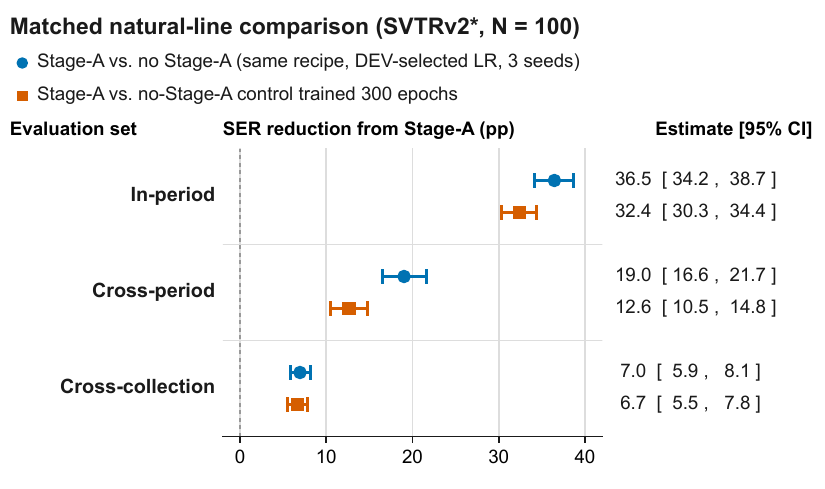}
\caption{\textbf{Stage-A effect on matched natural lines}, relative to the same-recipe no-Stage-A model and to the 300-epoch no-Stage-A control (reductions in SER, pp; paired tablet-bootstrap 95\% CIs).}
\label{fig:natural_matched}
\end{figure}

\paragraph{Matched protocol.} Each line is one rectangle cut from the tablet-side image: the union of
the line's annotated sign polygons plus a 15\% margin, with no cropping, resizing, or stitching of
individual signs. Hilprecht lines with at least two annotated signs are kept, every annotated token on them is a
target, and signs outside the inventory map to an explicit OOV class (751 outputs including blank). HT uses its
released line images and the references of its line manifest, which already omit the tokens outside both label
mappings (\S\ref{sec:crossperiod}); 419 of its 6{,}431 reference tokens are OOV. Both arms use the same 100
adaptation tablets, the same 48 DEV tablets with natural lines, three seeds, and the same 100-epoch
budget with patience 15. The learning rate is chosen per arm by seed-42 DEV \lcs from
$\{5{\times}10^{-5}, 10^{-4}, 3{\times}10^{-4}, 10^{-3}\}$ (selected:
$3{\times}10^{-4}$ with Stage-A, $10^{-4}$ without). Otherwise the arms differ only in
initialization: the current text-disjoint Stage-A checkpoint (with an added OOV row) versus random
initialization. Manifests were hashed before training. A 300-epoch no-Stage-A control uses the same
recipe. Natural-line sets keep more lines than the primary sets (906/1{,}729/1{,}408 lines, versus 670/1{,}243/1{,}320
primary lines); on the 670 primary in-period lines the reduction is 36.2\,pp [33.8, 38.6]. Stage-A lowers
SER in every domain, and the 300-epoch control remains 32.4\,pp (in-period) to
6.7\,pp (cross-collection) behind (Table~\ref{tab:natural_matched}, Figure~\ref{fig:natural_matched}).

\begin{table}[H]
\centering\revon\footnotesize\setlength{\tabcolsep}{3pt}
\caption{\textbf{Natural-line recognition with and without Stage-A} (\svtr, $N{=}100$, three seeds). SER (\%); reduction = no Stage-A $-$ Stage-A with a paired tablet-bootstrap 95\% CI. In-inventory scoring removes the OOV class from references and outputs.}
\label{tab:natural_matched}
\par\vspace{\baselineskip}
\begin{tabular*}{\linewidth}{@{\extracolsep{\fill}}lrcccc@{}}
\toprule
Set & Lines & no St.-A & no St.-A (300 ep.) & with St.-A & Reduction (pp) [95\% CI] \\
\midrule
In-period & 906 & 89.2 & 85.2 & \textbf{52.8} & 36.5 [34.2, 38.7] \\
In-period, strict & 291 & 88.8 & 86.2 & \textbf{62.4} & 26.4 [23.3, 29.4] \\
In-period, in-inventory scoring & 906 & 96.1 & 92.9 & \textbf{57.8} & 38.3 [35.9, 40.6] \\
Cross-period & 1{,}729 & 99.6 & 93.3 & \textbf{80.6} & 19.0 [16.6, 21.7] \\
Cross-collection (HT) & 1{,}408 & 99.2 & 98.9 & \textbf{92.2} & 7.0 [5.9, 8.1] \\
\bottomrule
\end{tabular*}
\end{table}

\paragraph{Line regions without annotation.} The same models also read lines located without any annotation
(Table~\ref{tab:natural_auto}): either the automatic system's predicted rows (the rows and order of the base
decoder, \S\ref{sec:pipeline}) or the raw, uncorrected output of an independent U-Net line segmenter. The
segmenter was trained on images of tablets from other periods (none Ur~III), none of them evaluation tablets
(checked by museum number, tablet--envelope twin, and CDLI text). Stage-A lowers whole-side SER by 27--35\,pp under every line
source and margin, and by 36--43\,pp when outputs tagged OOV are kept as errors.

\begin{table}[H]
\centering\revon\footnotesize\setlength{\tabcolsep}{3pt}
\caption{\textbf{Natural lines without line annotation at test time.} Whole-side SER (\%) of the natural-line \svtr models above, reading line crops from the automatic system's predicted rows or from an independent line segmenter, concatenated in predicted order. Main columns remove outputs tagged OOV before scoring, matching the whole-side reference, which excludes out-of-inventory regions; the last column keeps them under the full-output protocol of \S\ref{sec:metrics} (no Stage-A / Stage-A). CIs: paired tablet bootstrap in-period, CDLI-text bootstrap source-disjoint; 15\% line margin unless noted; sides in parentheses.}
\label{tab:natural_auto}
\par\vspace{\baselineskip}
\begin{tabular*}{\linewidth}{@{\extracolsep{\fill}}llcccc@{}}
\toprule
Line source & Set & no St.-A & with St.-A & Reduction (pp) [95\% CI] & OOV kept \\
\midrule
System's predicted rows & In-period (198) & 90.6 & \textbf{55.7} & 34.9 [32.7, 37.2] & 114.4 / 72.0 \\
System's predicted rows & Source-disjoint (34) & 92.6 & \textbf{58.2} & 34.4 [30.2, 39.0] & 118.9 / 80.7 \\
Independent segmenter & In-period (198) & 100.6 & \textbf{71.1} & 29.5 [27.4, 31.7] & 150.7 / 108.1 \\
\quad no margin & In-period (198) & 95.7 & \textbf{61.2} & 34.5 [32.1, 36.8] & 133.7 / 91.7 \\
Independent segmenter & Source-disjoint (34) & 101.2 & \textbf{74.5} & 26.7 [22.5, 31.0] & 152.2 / 116.2 \\
\quad no margin & Source-disjoint (34) & 95.9 & \textbf{62.2} & 33.7 [29.2, 38.0] & 134.0 / 97.8 \\
\bottomrule
\end{tabular*}
\end{table}

%% file: sections/development_checks.tex
\section{Development Checks}
\label{app:additional_checks}

These checks retain the primary checkpoint family and table entries. They examine the stopping condition of small-budget CRNN training and the DEV checkpoint and output-parsing sensitivity of the VLM. They do not reselect primary checkpoints by evaluation-set performance.

\subsection{CRNN small-budget control without early stopping}

We rerun the without-Stage-A $N{=}25$ \crnn for all three original seeds from random
initialization, using the same data, augmentation, learning rate $3\times10^{-4}$,
batch size 8, AdamW weight decay $10^{-4}$, and 100-epoch cosine schedule. Only patience
is extended from 15 to 100, so every run completes 100 epochs. Checkpoints still maximize
DEV LCS; none is selected using evaluation-set SER. The original failed-run records
remain unchanged in Table~\ref{tab:smallbudget}.

The selected epochs are 88, 71, and 89 for seeds 42, 123, and 456. The two formerly blank runs begin producing correct DEV signs after the old stopping
point. This shows that the epoch-15 failures were not permanent failures to learn.
However, the decoded models remain weak (Table~\ref{tab:crnn_fixed100}); longer training
does not rescue the small-budget scratch recognizer or overturn the pre-training benefit.

\begin{table}[htbp]
\centering\small
\caption{\textbf{Additional $N{=}25$ CRNN control.} Three-seed mean corpus SER (\%).
Original no-Stage-A includes the two declared empty-output failures; Fixed 100 decodes
three actual DEV-selected checkpoints. The With St.-A column gives the unchanged primary results.
The last column is Fixed 100 minus Stage-A, with paired-tablet 95\% CI.}
\label{tab:crnn_fixed100}
\par\vspace{\baselineskip}
\begin{tabular*}{\linewidth}{@{\extracolsep{\fill}}lcccc@{}}
\toprule
Domain & Original no St.-A & Fixed 100 & With St.-A & Gain [95\% CI] \\
\midrule
In-period & \Tnum{3.2}{100.51} & \Tnum{3.2}{99.44} & \Tnum{2.2}{33.63} & \Tci{2.2}{65.81}{63.41}{68.11} \\
Cross-period & \Tnum{3.2}{100.31} & \Tnum{3.2}{100.90} & \Tnum{2.2}{44.56} & \Tci{2.2}{56.35}{53.20}{59.00} \\
Cross-collection & \Tnum{3.2}{101.82} & \Tnum{3.2}{105.03} & \Tnum{2.2}{72.42} & \Tci{2.2}{32.61}{30.62}{34.73} \\
\bottomrule
\end{tabular*}
\end{table}

\subsection{VLM validation checkpoint and output-parsing checks}

The pilot compares the original DEV-selected $N{=}100$, seed-42 \vlm adapter
with steps 1000/2000/3000/4000 on 342 DEV lines (1,597 signs), fixing augmentation
(seed 12345), prompt, and deterministic generation, and saving all raw responses.
Unfiltered edit counts are 278/300/284/292/276. Step 4000 improves both SER
(17.41\% to 17.28\%, two errors) and LCS. This does not show conflicting metric
rankings: training-time validation used changing augmentation draws.

The historical \vlm evaluator normalizes sign names and removes tokens outside the
inventory before scoring. Its saved-token SER is not unfiltered generation SER.
For these five DEV checkpoints, explicitly retaining unknown output tokens leaves
the total edit counts unchanged, although some individual alignments differ. This
check is limited to these five DEV checkpoints. Original oracle-line
results retain their saved-token scoring protocol.

The symmetric check covers three seeds with/without Stage-A
(Table~\ref{tab:vlm_dev_symmetric}), using identical images, references, prompts,
and generation settings. Scoring normalizes known signs but retains unknown tokens
and response prefixes. Seed-42 Stage-A matches the pilot. The latest archived
adapters are step 3000 for Stage-A seed 456 and 4000 otherwise, not necessarily
final epochs.

\begin{table}[H]
\centering\small
\caption{\textbf{Symmetric VLM DEV checkpoint check.} Fixed identical inputs, $N{=}100$, three seeds per arm. ``Selected'' is the original DEV-LCS checkpoint; ``later'' is the latest archived periodic adapter (step 3000 for Stage-A seed 456, 4000 otherwise). Both SER columns retain unknown output tokens; $\Delta E$ is later minus selected edit count. This is DEV sensitivity, not a new test result.}
\label{tab:vlm_dev_symmetric}
\par\vspace{\baselineskip}
\begin{tabular*}{\linewidth}{@{\extracolsep{\fill}}lcccc@{}}
\toprule
Stage-A & Seed & Selected SER (\%) & Later SER (\%) & $\Delta E$ \\
\midrule
Yes & \Tnum{3.0}{42} & \Tnum{2.2}{17.41} & \Tnum{2.2}{17.28} & \Tnum{+1.0}{-2} \\
Yes & \Tnum{3.0}{123} & \Tnum{2.2}{17.16} & \Tnum{2.2}{16.84} & \Tnum{+1.0}{-5} \\
Yes & \Tnum{3.0}{456} & \Tnum{2.2}{17.60} & \Tnum{2.2}{17.72} & \Tnum{+1.0}{+2} \\
No & \Tnum{3.0}{42} & \Tnum{2.2}{41.58} & \Tnum{2.2}{41.70} & \Tnum{+1.0}{+2} \\
No & \Tnum{3.0}{123} & \Tnum{2.2}{43.64} & \Tnum{2.2}{43.77} & \Tnum{+1.0}{+2} \\
No & \Tnum{3.0}{456} & \Tnum{2.2}{42.20} & \Tnum{2.2}{42.52} & \Tnum{+1.0}{+5} \\
\bottomrule
\end{tabular*}
\end{table}

Changes are two to five errors over 1,597 signs. Unfiltered SER and LCS agree
for all six pairs; with the historical parser, without-Stage-A seed 123 has one
fewer error but lower LCS at the later point. Filtering changes at most three
errors (0.19\,pp) across these twelve DEV outputs. These DEV checks leave the primary checkpoints and three-domain results unchanged.

%% file: sections/added_appendix.tex
\FloatBarrier
\section{Independent Stage-A Seeds for the Attribution Contrast}
\label{app:stageaseeds}
We retrained the attested and within-split shuffled Stage-A models of \S\ref{sec:strict} with two further
seeds (123, 456). The seed changes initialization, data order, the train/validation split, and the shuffle
permutation; corpus, token exposure, update budget, and renderer are unchanged. Each checkpoint then follows
the locked Stage-B protocol (DEV-selected learning rate, three Stage-B seeds). On all in-period lines every seed's
95\% CI lies above zero, and the across-seed 95\% $t$-interval is [3.09, 6.19]\,pp; on the strict
in-period subset the gain is positive for every seed (Table~\ref{tab:stageaseeds}). Across the three Stage-A seeds, the
shuffled model's strict-subset SER varies by at most 2.1, 5.7, and 4.5\,pp
(in-period, cross-period, cross-collection), far less than the 45.1, 46.2, and 23.1\,pp line-format step of
Figure~\ref{fig:strict}.

\begin{table}[H]
\centering\revon\footnotesize\setlength{\tabcolsep}{3pt}
\caption{\textbf{Attested vs.\ shuffled Stage-A across independent pre-training seeds} (\svtr, $N{=}100$). Gain from attested order (shuffled $-$ attested SER, pp; positive = attested helps), each with a paired tablet-bootstrap 95\% CI over its three Stage-B seeds. Seed 42 reproduces Figure~\ref{fig:strict}; its all-lines CIs come from a separate bootstrap draw (Table~\ref{tab:ladder_panelb} differs by at most 0.1).}
\label{tab:stageaseeds}
\par\vspace{\baselineskip}
\begin{tabular*}{\linewidth}{@{\extracolsep{\fill}}lcccc@{}}
\toprule
Set & Seed 42 (Fig.~\ref{fig:strict}) & Seed 123 & Seed 456 & Mean \\
\midrule
In-period, all lines & 4.41 [3.29, 5.55] & 5.35 [4.01, 6.67] & 4.16 [2.79, 5.55] & 4.64 \\
In-period, strict & 1.16 [$-$0.68, 2.98] & 3.87 [2.15, 5.54] & 1.55 [$-$0.59, 3.82] & 2.19 \\
Cross-period, all lines & 4.20 [2.99, 5.27] & 2.77 [1.91, 3.65] & $-$1.34 [$-$2.63, $-$0.19] & 1.88 \\
Cross-period, strict & 3.54 [2.15, 4.77] & 2.21 [1.21, 3.22] & $-$1.91 [$-$3.22, $-$0.75] & 1.28 \\
Cross-collection, all lines & 1.72 [0.77, 2.67] & 2.37 [1.63, 3.09] & $-$0.96 [$-$1.84, $-$0.10] & 1.04 \\
Cross-collection, strict & 1.80 [0.85, 2.76] & 2.26 [1.55, 2.98] & $-$1.02 [$-$1.91, $-$0.14] & 1.01 \\
\bottomrule
\end{tabular*}
\end{table}

\FloatBarrier
\section{Source-Disjoint Whole-Side Evaluation and Verifier Attribution}
\label{app:sourcedisjoint}
\paragraph{Set.} From HeiCuBeDa \citep{mara2019heicubeda} we selected Ur~III objects, each with a CDLI transliteration, whose images
appear nowhere in our data: in no detector, classifier, or recognizer training or adaptation split, in no DEV, test,
or HT list, and in no source of the isolated-sign dictionary. Objects were matched by museum number, tablet--envelope
twin, and CDLI text. After excluding fragments, unclear orientations, incomplete transliterations, and three texts whose objects or tablet--envelope twins contribute dictionary crops, 34 sides from 18
tablets and envelopes remain (13 texts, 674 reference signs). References were built from the CDLI transliterations and
hashed before any model run. The six-gram LM was rebuilt without these texts; rebuilding with the original exclusion
list reproduces the deployed LM byte for byte. Of the 387{,}315 Stage-A lines, 100 share
text with these tablets; those lines are rendered from other tablets' crops. Transliteration references list more signs than
box annotations (on in-period data they raise the same system's SER by 5.8\,pp), so absolute values are not comparable
across columns of Table~\ref{tab:sourcedisjoint}.

\begin{figure}[tbp]
\centering\revon
\includegraphics[width=0.82\linewidth]{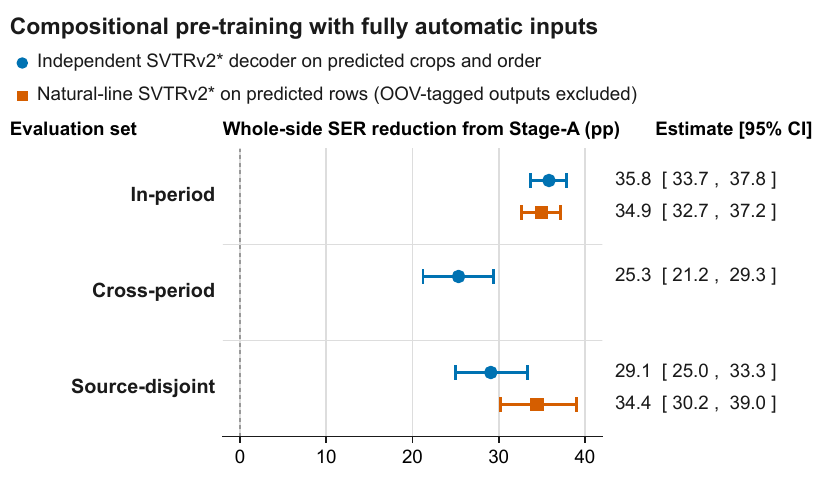}
\caption{\textbf{Stage-A effect with fully automatic inputs}: independent \svtr on predicted crops and order, and natural-line \svtr on predicted rows (outputs tagged OOV excluded). Sides: 198 in-period, 287 cross-period, 34 source-disjoint (transliteration references).}
\label{fig:auto_inputs}
\end{figure}

\paragraph{Verifier and decoder with and without Stage-A.} The no-Stage-A verifier is the matched $N{=}100$ \svtr of
Table~\ref{tab:main}. For each recognizer seed, $\lambda$ is chosen on DEV from the grid of
Appendix~\ref{app:svtr_verifier} with the historical objective, identically for both arms; test data are never used.
In-period, each of the three runs uses its own classifier seed and DEV candidate pool; the cross-period and
source-disjoint sets use the classifier-123 system and its DEV pool. The cross-period sides were seen by the detector
(Appendix~\ref{app:auto_refinement}). With Stage-A the verifier improves every system; without it, the verifier is assigned
weight 0 in two of three in-period runs and in every other run (four of the five distinct DEV selections). At the frozen
weight for both verifiers, Stage-A still gives lower SER in-period and cross-period; on the 34 source-disjoint sides that
interval includes zero. The independent decoder's SER rises from
46.39/68.29/56.68\% to 82.22/93.63/85.76\% without Stage-A (Table~\ref{tab:sourcedisjoint}, Figure~\ref{fig:auto_inputs}).
With the six-gram LM in the decoder, the Stage-A \svtr reaches 43.88\% in-period, versus 76.54\% without
Stage-A and 74.38\% for the 300-epoch no-Stage-A control (30.50 [28.42, 32.46]\,pp).
More adaptation data does not close this gap: with the classifier's 214 adaptation tablets instead of 100, a no-Stage-A \svtr reaches 53--57\%
DEV \lcs, versus 85--86\% with Stage-A.

\begin{table}[H]
\centering\revon\footnotesize\setlength{\tabcolsep}{3pt}
\caption{\textbf{Whole-side SER (\%) with and without Stage-A in the sequence model.} Means of three runs; within each run, all rows share the detector, classifier, candidate pool, layout, LM, and references. DEV-selected $\lambda$ uses the same grid (including 0) for both verifiers; $\lambda{=}0.25$ is the full system's frozen weight. The LM row decodes in-period with the system's six-gram LM, its weights chosen on DEV for each model. The source-disjoint column uses transliteration references and a CDLI-text bootstrap; in-period and cross-period use tablet bootstraps. Sides in parentheses.}
\label{tab:sourcedisjoint}
\par\vspace{\baselineskip}
\begin{tabular*}{\linewidth}{@{\extracolsep{\fill}}lccc@{}}
\toprule
 & In-period (198) & Cross-period (287) & Source-disjoint (34) \\
\midrule
Base system, no verifier & 39.88 & 60.72 & 48.66 \\
Full system (Stage-A verifier, $\lambda{=}0.25$) & 38.05 & 58.09 & 46.49 \\
\quad reduction vs.\ base [95\% CI] & 1.83 [1.15, 2.53] & 2.63 [1.91, 3.44] & 2.18 [0.64, 3.66] \\
Stage-A verifier, DEV-selected $\lambda$ & 38.09 & 58.81 & 47.38 \\
No-Stage-A verifier, DEV-selected $\lambda$ & 39.57 & 60.72 & 48.66 \\
\quad Stage-A vs.\ no Stage-A [95\% CI] & 1.48 [0.95, 2.05] & 1.91 [1.41, 2.48] & 1.29 [0.13, 2.32] \\
No-Stage-A verifier, $\lambda{=}0.25$ & 39.33 & 59.45 & 47.48 \\
\quad Stage-A vs.\ no Stage-A at $\lambda{=}0.25$ [95\% CI] & 1.28 [0.66, 1.92] & 1.36 [0.54, 2.20] & 0.99 [$-$0.58, 2.59] \\
Independent \svtr: Stage-A / no Stage-A & 46.39 / 82.22 & 68.29 / 93.63 & 56.68 / 85.76 \\
\quad with the six-gram LM & 43.88 / 76.54 & --- & --- \\
\bottomrule
\end{tabular*}
\end{table}

\section{Combining the Per-Sign Classifier with the Sequence Model}
\label{app:persign_fusion}
For each line with GT sign boxes, the $K{=}64$ best label sequences from DeiT3's top-5 labels per position (plus the
\svtr greedy output when its length matches) are rescored by $\sum_t \log p_{\mathrm{DeiT3}} + w \log P_{\mathrm{CTC}}$, where
the CTC probability is computed by \svtr on the stitched line. The weight $w$ is chosen on DEV for each seed (DeiT3 and \svtr checkpoints of the same seed) from
$\{0, 0.05, 0.1, 0.2, 0.3, 0.5, 0.75, 1, 1.5, 2, 3\}$. Rescoring with the Stage-A \svtr beats DeiT3 alone on all three sets; with the no-Stage-A
\svtr the off-domain gains disappear (Table~\ref{tab:persign_fusion}). On strict subsets, rescoring lowers DeiT3's SER from
22.2\% to 17.0\% in-period and from 28.3\% to 26.1\% cross-period. Compared directly, the Stage-A combination
is better than the no-Stage-A one by 4.2 [3.3, 5.3]\,pp in-period and
3.0 [2.3, 3.6]\,pp cross-period.

\begin{table}[H]
\centering\revon\footnotesize\setlength{\tabcolsep}{3pt}
\caption{\textbf{Combining the per-sign classifier with the sequence model} (GT sign boxes, $N{=}100$, three seeds). SER (\%). Reduction = DeiT3 $-$ combination with the Stage-A \svtr, paired tablet-bootstrap 95\% CI.}
\label{tab:persign_fusion}
\par\vspace{\baselineskip}
\begin{tabular*}{\linewidth}{@{\extracolsep{\fill}}lccccc@{}}
\toprule
Set & DeiT3 & \svtr & DeiT3+\svtr & Reduction (pp) & DeiT3+\svtr (no St.-A) \\
\midrule
In-period & 21.0 & 21.5 & \textbf{14.6} & 6.46 [5.43, 7.58] & 18.8 \\
Cross-period & 27.8 & 38.4 & \textbf{25.1} & 2.68 [1.95, 3.47] & 28.1 \\
Cross-collection (HT) & 62.4 & 71.2 & \textbf{61.7} & 0.71 [0.02, 1.38] & 62.4 \\
\bottomrule
\end{tabular*}
\end{table}

\FloatBarrier
\section{Inference Cost}
\label{app:speed}
On the same 670 lines (Table~\ref{tab:speed}), \svtr has 11$\times$ fewer parameters, needs 28$\times$ fewer FLOPs,
uses less peak GPU memory, and runs 12$\times$ faster on CPU and 1.4$\times$ faster on a GPU when lines are read one at a
time; it also reads a line without a box for every sign. DeiT3 is faster only on a GPU with crops pooled across lines.

\begin{table}[H]
\centering\revon\footnotesize\setlength{\tabcolsep}{3pt}
\caption{\textbf{Inference cost} on the 670 in-period lines (RTX~5090, fp32; recognizer only). \svtr reads a whole line; DeiT3 classifies each GT crop. Peak memory is allocated memory with each model alone on the GPU, weights included. Our Stage-B \svtr passes no attention padding mask (Appendix~\ref{app:recognizer_recipe}), so padded GPU batches change its outputs; its unpadded single-line timing is reported.}
\label{tab:speed}
\par\vspace{\baselineskip}
\begin{tabular*}{\linewidth}{@{\extracolsep{\fill}}lcc@{}}
\toprule
 & \svtr & DeiT3-medium \\
\midrule
Parameters (M) & 3.5 & 38.7 \\
GFLOPs per line (5 signs) & 2.8 & 79.9 \\
CPU, 8 threads, median ms per line (first 100 lines) & 9.6 & 118.4 \\
GPU, one line per forward, total s (670 lines) & 2.06 & 2.80 \\
GPU, crops pooled across lines (batch 256), total s & --- & 1.14 \\
Peak GPU memory, one line per forward (MiB) & 189 & 228 \\
\bottomrule
\end{tabular*}
\end{table}

\FloatBarrier
\section{Common CTC/VLM Subset}
\label{app:commonsubset}
On the lines whose references are token-identical under both label mappings, all models are scored against the same
sequences (Table~\ref{tab:commonsubset}). The Stage-A removal costs are within 1.3\,pp of Table~\ref{tab:main}.

\begin{table}[H]
\centering\revon\footnotesize\setlength{\tabcolsep}{3pt}
\caption{\textbf{Stage-A removal cost on a common CTC/VLM line subset} with token-identical references (number of lines in parentheses); SER increase (pp) with paired tablet-bootstrap 95\% CIs. VLM generations were filtered to the inventory when first stored, as in Table~\ref{tab:main}.}
\label{tab:commonsubset}
\par\vspace{\baselineskip}
\begin{tabular*}{\linewidth}{@{\extracolsep{\fill}}lccc@{}}
\toprule
Set (lines) & \svtr & \crnn & \vlm \\
\midrule
In-period (572) & 49.6 [47.3, 51.9] & 47.2 [44.8, 49.5] & 26.6 [24.6, 28.7] \\
Cross-period (1{,}155) & 51.7 [48.7, 54.3] & 48.2 [45.3, 50.6] & 35.0 [32.5, 37.0] \\
Cross-collection (1{,}056) & 26.9 [25.0, 28.9] & 24.3 [22.5, 26.2] & 24.9 [23.1, 26.8] \\
\bottomrule
\end{tabular*}
\end{table}

\FloatBarrier
\section{Length and Alignment in the Training-Format Contrast}
\label{app:length}
Part of the gap between singleton-matched and shuffled-line pre-training reflects output length: the singleton model
under-emits on multi-sign lines. Where both models output exactly the reference length, the line-format model is still
far more accurate per position, and the gap is already large on 3-sign lines (Table~\ref{tab:length}). Equal output
length controls this inference-time difference; it does not isolate the CTC alignment learned in training.

\begin{table}[H]
\centering\revon\footnotesize\setlength{\tabcolsep}{3pt}
\caption{\textbf{Length and alignment within the training-format contrast} (all lines, $N{=}100$). Deletion rate (\%) and output/reference length for singleton-matched vs.\ shuffled-line pre-training; per-position accuracy (\%; singleton / lines) on lines where both models output the reference length (312/244/177 line--seed pairs); and the SER gap (singleton $-$ lines, paired tablet-bootstrap 95\% CI) on 3-sign lines.}
\label{tab:length}
\par\vspace{\baselineskip}
\begin{tabular*}{\linewidth}{@{\extracolsep{\fill}}lcccccc@{}}
\toprule
Set & Del.\ sing. & Del.\ lines & Len.\ sing. & Len.\ lines & Aligned acc. & 3-sign gap (pp) \\
\midrule
In-period & 41.4 & 0.8 & 0.61 & 0.99 & 63.8 / \textbf{86.7} & 33.4 [30.9, 35.9] \\
Cross-period & 53.5 & 0.4 & 0.47 & 1.00 & 23.8 / \textbf{64.7} & 37.2 [34.0, 40.3] \\
Cross-collection (HT) & 56.3 & 0.7 & 0.44 & 1.00 & 11.9 / \textbf{31.9} & 19.7 [17.6, 21.8] \\
\bottomrule
\end{tabular*}
\end{table}